\documentclass[11pt]{article}
\usepackage{adjustbox}
\usepackage[final]{acl}

\usepackage{times}
\usepackage{latexsym}

\usepackage{booktabs}
\usepackage{tabularx}
\usepackage{ragged2e}
\usepackage{array}

\usepackage[T1]{fontenc}
\usepackage[utf8]{inputenc}

\usepackage{microtype}

\usepackage{inconsolata}

\usepackage{graphicx}
\usepackage{booktabs}
\usepackage{multirow}
\usepackage{makecell}
\usepackage{amsmath}
\usepackage{amssymb}

\usepackage{hyperref}

\title{PRISP: Privacy-Safe Few-Shot Personalization via Lightweight Adaptation}

\author{
Junho Park$^{1}$\thanks{Equal contribution} \qquad
    Dohoon Kim$^{1}$\footnotemark[1] \qquad
    Taesup Moon$^{1,2}$\thanks{Corresponding author}
    \\
    $^{1}$ Department of Electrical and Computer Engineering, Seoul National University \\
    $^{2}$ ASRI/INMC/IPAI/AIIS, Seoul National University \\
    {\tt\small \{bryan123, dohoon.kim, tsmoon\}@snu.ac.kr }
}

\begin{document}
\maketitle
\begin{abstract}
Large language model (LLM) personalization aims to adapt general-purpose models to individual users. Most existing methods, however, are developed under data-rich and resource-abundant settings, often incurring privacy risks. In contrast, realistic personalization typically occurs after deployment under (i) extremely limited user data, (ii) constrained computational resources, and (iii) strict privacy requirements. We propose PRISP (\textbf{Pri}vacy-\textbf{S}afe Few-Shot \textbf{P}ersonalization), a lightweight personalization framework tailored to these constraints. PRISP leverages a Text-to-LoRA hypernetwork to generate task-aware LoRA parameters from task descriptions, and enables efficient user personalization by optimizing a small subset of task-aware LoRA parameters together with minimal additional modules using few-shot user data. Experiments on a few-shot variant of the LaMP benchmark demonstrate that PRISP achieves strong overall performance compared to prior approaches, while reducing computational overhead and eliminating privacy risks. Code is available at \href{https://github.com/kingthebryan/PRISP}{\tt{https://github.com/kingthebryan/PRISP}}.
\end{abstract}

\section{Introduction}

Large Language Models (LLMs) \citep{gpt3, touvron2023llama, team2024gemma, yang2025qwen3} have demonstrated remarkable generalization across diverse natural language tasks. However, a \textit{one-size-fits-all} approach often fails to capture the specific preferences, stylistic nuances, and evolving goals of individual users. This limitation has sparked growing interest in \textit{personalization} \citep{Salemi2024LaMP, chen2024largemeetperson, tan2024democratizing, zhang2024personalizationsurvey}, emphasizing the need for model adaptation tailored to specific user contexts.

Most existing research on personalization has been developed and evaluated under data-rich settings
\citep{Salemi2024LaMP, kumar2024longlamp, salemi2025lampqa, liu2025survey}.
In particular, LaMP \citep{Salemi2024LaMP}, one of the most widely adopted benchmarks, is constructed with substantial amounts of user data,
and current personalization methods are primarily validated under these favorable conditions
\citep{tan2024personalized, tan2024democratizing, kim2025personalized},
often achieving better performance than prompt-based methods that rely on retrieval
\citep{lewis2020retrieval, izacard2022fewretrieval, Richardson2023Integrating, mysore2023pearl}.

\begin{figure}
    \centering
    \includegraphics[width=\columnwidth]{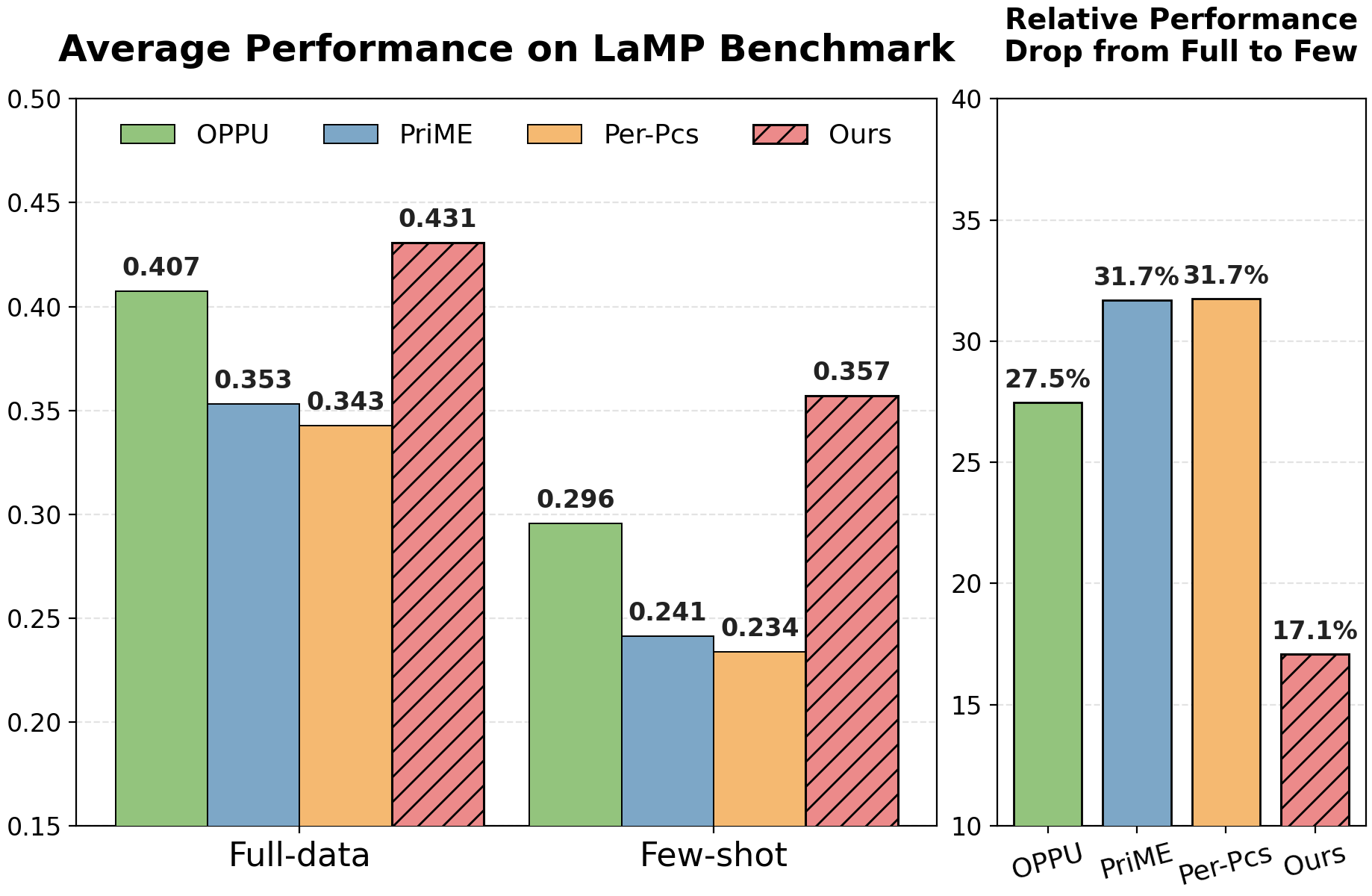}
    \caption{When personalization relies on less than 1\% of the full training data, existing methods exhibit substantial performance degradation, highlighting the challenge of robust few-shot personalization. The exact few-shot setting is described later in Table~\ref{tab:fewshot-scenario}.}

    \label{fig:full_to_few}
\end{figure}

However, realistic personalization typically occurs after deployment and relies on a small amount of user data, rather than large-scale data as in standard language model training \citep{zhang2024personalizationsurvey, kim-yang-2025-shot}. In practice, it is natural to expect that each user provides only a few task-relevant examples. Accordingly, we modify the LaMP benchmark to emphasize few-shot personalization by reducing the available data, rather than relying on the original full-data setting. As shown in Figure~\ref{fig:full_to_few}, existing methods rely heavily on abundant data and often fail when operating in the few-shot regime.

\begin{figure*}[t]
    \centering
    \includegraphics[width=\textwidth]{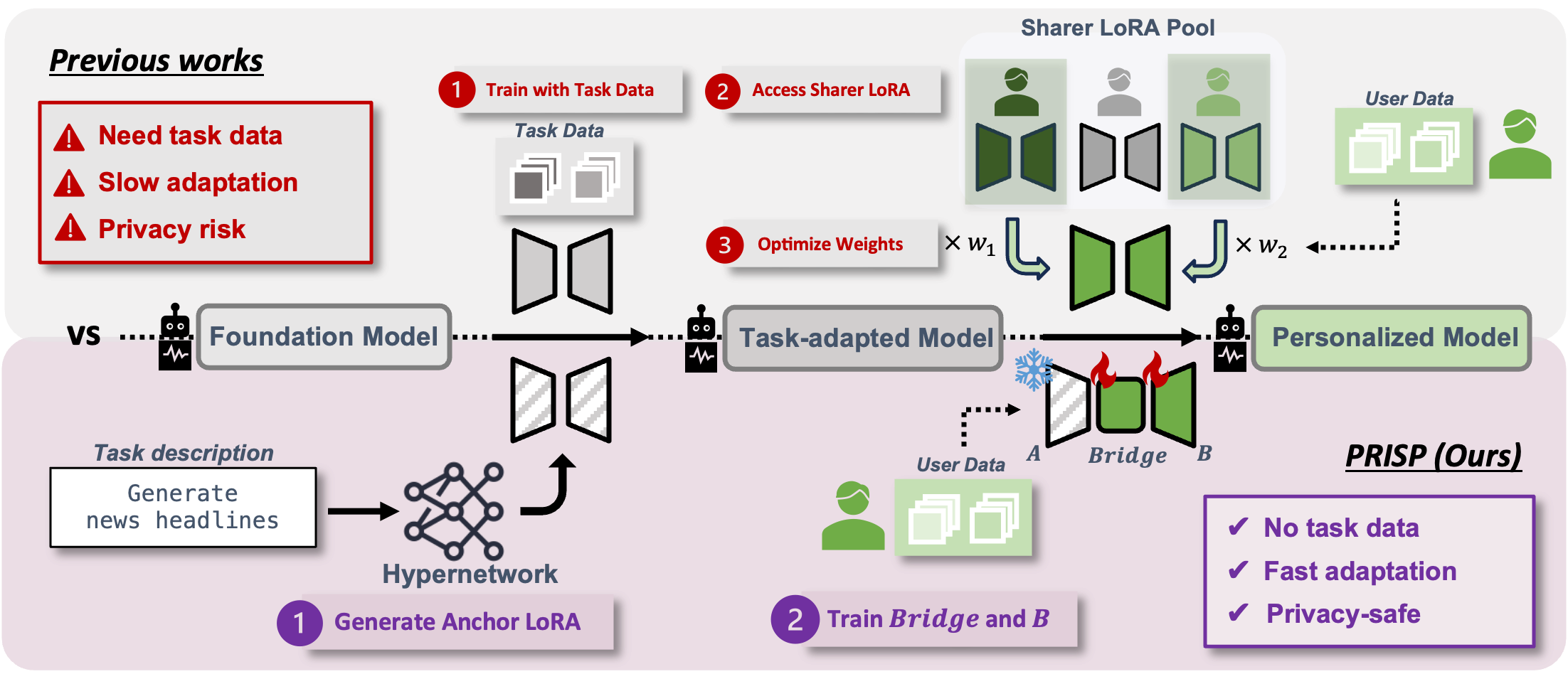}
    \caption{Overview of PRISP in comparison to previous works. \textbf{Top:} Prior methods rely on task data and other users’ parameters (sharer LoRAs) for personalization. \textbf{Bottom:} PRISP generates a task-aware anchor LoRA from a task description without task data via a hypernetwork, and personalizes the model by training only a lightweight \textit{bridge} module and the $B$ matrix. This enables task-data-free, privacy-safe, and fast personalization.}
    \label{fig:Main-figure}
\end{figure*}

Beyond data scarcity, personalization must also account for resource constraints. Since personalization is driven by private user-specific data, it is often desirable and necessary for adaptation to be performed on edge devices \citep{liu2025survey}. This setting imposes strict limits on computational resources, such as GPU memory and training time. While prior work \citep{tan2024personalized, kim2025personalized} seeks to reduce GPU costs by relying less on gradient-based training through parameter merging, these methods still incur non-trivial training-time overhead from optimization procedures, limiting their practicality in realistic edge scenarios. This highlights the need for personalization methods that are not only data-efficient but also computationally lightweight.

Furthermore, privacy is a fundamental concern in personalization, as user-specific data and adapted parameters may encode sensitive information \citep{liu2025survey}. However, many existing approaches \citep{kim2025personalized, tan2024personalized, zhang2025proper} rely on reusing parameters from other users to leverage inter-user similarity, as illustrated in Figure~\ref{fig:Main-figure}. This implicitly assumes that sharing personalized parameters across users is acceptable. Such designs introduce a fundamental privacy risk, as personalized parameters may encode information derived from other users’ private data~\citep{carlini2023quantifying, inan2021trainingdataleakage, nasr2025scalable}.

In this paper, we propose PRISP (\textbf{Pri}vacy-\textbf{S}afe Few-Shot \textbf{P}ersonalization), a personalization framework designed to operate under limited data, constrained computational resources, and strict privacy guarantees. Our approach leverages a Text-to-LoRA \citep{charakorn2025texttolora} hypernetwork that generates task-aware LoRA \citep{Hu2022LoRA} parameters directly from natural language task descriptions. 
To enable efficient and effective personalization with minimal user data, we freeze the input-side LoRA parameters and introduce lightweight modules that increase model flexibility. These modules, together with the output-side LoRA parameters, are trained using few-shot user data. 
% This design enables efficient on-device personalization while avoiding heavy computational costs and eliminating privacy risks associated with sharing user-specific information.

Our contributions are summarized as follows:
\begin{itemize}

    \item We propose PRISP, a personalization framework designed for realistic settings with limited user data, constrained computational resources, and strict privacy guarantees.
    
    \item Extensive experiments show that PRISP achieves strong performance in both few-shot and full-data personalization, with particularly robust gains in the few-shot regime, while reducing overall computational costs and eliminating privacy risks from user-specific data and parameter sharing.

    % \item Extensive experiments show that PRISP achieves strong performance in \textbf{few-shot} personalization, while reducing \textbf{computational costs}, and eliminating \textbf{privacy risks} from parameter sharing and task-specific data.

    \item Through extensive ablation studies and analyses, we validate the effectiveness of our design choices, demonstrating consistent few-shot performance across data scales and strong adaptability to unseen tasks.
\end{itemize}

\section{Related Work}

\paragraph{Parameter-Efficient Fine-Tuning.}
Parameter-efficient fine-tuning (PEFT) \citep{he2022towards} adapts large language models using a small number of additional parameters. While various PEFT approaches such as adapters \citep{Houlsby2019Parameter, pfeiffer2020madadpaterP}, prefix tuning \citep{li2021prefix}, and prompt tuning \citep{lester2021prompt} have been explored, Low-Rank Adaptation (LoRA) \citep{Hu2022LoRA} has been widely adopted for its simplicity and effectiveness, enabling efficient adaptation through low-rank weight updates.

\paragraph{Personalization.}Personalization aims to tailor models to individual users’ preferences \citep{liu2025survey}. Retrieval-Augmented Generation (RAG) \citep{Salemi2024LaMP} retrieves user-specific historical interactions, while Profile-Augmented Generation (PAG) \citep{Richardson2023Integrating} summarizes a user’s history into a compact profile representation; in both cases, the retrieved information is merged into the input prompt to guide personalized generation. 
% PLoRA \citep{Zhang2024PLORA} employs a shared LoRA backbone conditioned on user embeddings to enable user-aware adaptation without training separate adapters.
OPPU \citep{tan2024democratizing} directly fine-tunes an independent LoRA module for each user based on a task-adapted model. Per-Pcs \citep{tan2024personalized} instead forms a personalized adapter by selecting and merging multiple shared anchor modules, enabling user-specific adaptation without training a fully separate LoRA from scratch. PriME \citep{kim2025personalized} further addresses privacy concerns through a privacy-balanced objective optimized with gradient-free evolutionary strategies.

\section{Method}
\subsection{Preliminaries}
\paragraph{Low-Rank Adaptation \citep{Hu2022LoRA}.} LoRA enables parameter-efficient adaptation of large language models by augmenting frozen pre-trained weights with low-rank updates.
Specifically, for a given layer $l$ with pre-trained weights $W^{l} \in \mathbb{R}^{d_{\text{out}}^{l} \times d_{\text{in}}^{l}}$, LoRA parameterizes the weight update as
\begin{equation}
\Delta W^{l} = B^{l} A^{l},
\notag
\end{equation}
where $A^{l} \in \mathbb{R}^{r \times d_{\text{in}}^{l}}$ and
$B^{l} \in \mathbb{R}^{d_{\text{out}}^{l} \times r}$ are trainable low-rank matrices with rank $r \ll \min(d_{\text{in}}^{l}, d_{\text{out}}^{l})$.
The adapted weight is then given by
\begin{equation}
\tilde{W}^{l} = W^{l} + B^{l} A^{l}
\notag
\end{equation}

\paragraph{Text-to-LoRA \citep{charakorn2025texttolora}.} Building on this formulation, Text-to-LoRA introduces a hypernetwork-based approach that directly generates LoRA parameters $(A^{l}, B^{l})$ from natural language task descriptions, enabling task adaptation without gradient-based fine-tuning.
Formally, given a task description $t$, a hypernetwork $H_{\psi}(\cdot)$ produces layer-wise LoRA parameters for a set of $L$ target layers:
\begin{equation}
\{(A_t^{l}, B_t^{l})\}_{l=1}^{L} = H_{\psi}(t).
\notag
\end{equation}

% \noindent\textbf{DoMIX~\citep{kim2025domix}. }
% DoMIX introduces a bridge matrix that enables structured mixing among multiple LoRA modules while improving data efficiency. 
% Specifically, the low-rank factors from $K$ LoRA modules are concatenated as
% \begin{equation}
% B_{\text{cat}} =
% \big[ B^{(1)}, \ldots, B^{(K)} \big], \quad
% A_{\text{cat}} =
% \begin{bmatrix}
% A^{(1)} \\
% \vdots \\
% A^{(K)}
% \end{bmatrix},
% \end{equation}
% where $B_{\text{cat}} \in \mathbb{R}^{d_{\text{out}} \times Kr}$ and
% $A_{\text{cat}} \in \mathbb{R}^{Kr \times d_{\text{in}}}$.

% DoMIX defines the composed update as
% \begin{equation}
% \Delta W_{\text{DoMIX}} = B_{\text{cat}} \, P \, A_{\text{cat}},
% \end{equation}
% where $P \in \mathbb{R}^{Kr \times Kr}$ is a learnable bridge matrix that governs interactions across different LoRA components.

% Importantly, DoMIX freezes the concatenated $B_{\text{cat}}$ matrices during training and only learns the bridge matrix $P$ (and optionally $A_{\text{cat}}$). 
% By restricting optimization to a small set of mixing parameters, DoMIX significantly reduces the number of trainable parameters, thereby improving data efficiency.
% This design is particularly beneficial in few-shot personalization settings, where only limited user-specific data is available.

% The adapted model parameters are then given by
% \begin{equation}
% W' = W + \Delta W_{\text{DoMIX}}.
% \end{equation}

\subsection{PRISP}
Our method, PRISP, extends the Text-to-LoRA paradigm to user-level personalization.
The framework is composed of two stages:
(i) constructing a task-aware \emph{anchor LoRA} from a natural language task description, and
(ii) adapting this anchor to individual users using limited user data.
An overview of the framework is shown in Figure~\ref{fig:Main-figure}.

\paragraph{Stage~1: Anchor LoRA Construction.}
The goal of Stage~1 is to initiate personalization by constructing a task-aware LoRA module that serves as a reasonable anchor for subsequent user adaptation.
Rather than learning task knowledge from large amounts of task-specific data, we leverage a pretrained Text-to-LoRA hypernetwork to generate an initial set of task-aware LoRA parameters directly from a natural language task description.

Given a task description $t$, the hypernetwork $H_\psi$ produces a set of layer-wise LoRA parameters
$\{(A_t^l, B_t^l)\}_{l=1}^{L}$.
These parameters are instantiated as a task-aware LoRA module and attached to the frozen backbone weights $W^l$.
At each layer $l$, this results in the following update:
\begin{equation}
\Delta W_t^l = B_t^l A_t^l,
\quad l = 1,\dots,L.
\notag
\end{equation}
We refer to this task-level LoRA module as the \emph{anchor LoRA}.

\paragraph{Stage~2: User-level Personalization.}
Starting from the anchor LoRA obtained in Stage~1, Stage~2 adapts the model to an individual user using limited user data.
In contrast to One-PEFT-Per-User (OPPU)~\citep{tan2024democratizing}, which introduces a separate LoRA module for each user, our approach performs personalization by refining the anchor LoRA, leading to improved data efficiency and stability in few-shot settings.

To this end, we introduce a lightweight, learnable bridge matrix $C^l$ for each LoRA layer.
The user-specific weight update at layer $l$ is parameterized as
\begin{equation}
\Delta W_{\text{user}}^l
= B_t^l \, C^l \, A_t^l,
\quad l = 1,\dots,L.
\notag
\end{equation}
Each $C^l$ is initialized as the identity matrix $I$, ensuring that the initial user model exactly recovers the anchor LoRA.
During user personalization, the anchor input matrices $A_t^l$ are frozen to stabilize adaptation under few-shot constraints, while $C^l$ and $B_t^l$ are trained using the user’s interaction history $\mathcal{H}_{\text{user}}$ by minimizing
\begin{equation}
\mathbb{E}_{(x,y)\sim\mathcal{H}_{\text{user}}}
\left[
\mathcal{L}\big(f(x; W_{\text{user}}), y\big)
\right].
\notag
\end{equation}
Here, $\mathcal{L}(\cdot,\cdot)$ denotes the task loss,
and $f(x; W_{\text{user}})$ is the personalized model with user-specific weights $W_{\text{user}}$.

Overall, Stage~2 enables robust few-shot personalization by introducing minimal and well-constrained user-specific updates on top of the anchor LoRA.
Our design is inspired by DoMIX~\citep{kim2025domix}, but is developed for a different setting of user-level personalization under few-shot constraints.

\section{Experiments}

\subsection{Experimental Settings}

\paragraph{Benchmark and Models.} We use the LaMP  benchmark~\citep{Salemi2024LaMP}, which consists of six public personalization tasks, including three text classification tasks and three text generation tasks. Additional benchmark details are provided in Appendix~\ref{sec:appendix-datasets}.

We adopt Qwen3-0.6B \citep{yang2025qwen3} as the base model, as its compact architecture is well suited for realistic personalization scenarios where computational resources are limited.
In addition, to assess performance in more resource-abundant settings, we report results on a larger backbone, Llama-3.1-8B-Instruct \citep{grattafiori2024llama3}, in Appendix \ref{appendix:C_addditional Analyses}.

\paragraph{Few-Shot Scenario.}
In few-shot scenarios, for all methods excluding ours, task adapting stages are conducted using 50 random samples from the full task data, while the user-level personalization stage uses the 10 most recent history items per user. This corresponds to less than 1\% of the full task data and approximately 10\% of each target user’s available history shown in Table \ref{tab:fewshot-scenario}.

In contrast, PRISP does not use \textit{any} task data. Instead, it initializes personalization with a task-aware anchor LoRA generated via hypernetwork and then performs user-level personalization using 10 target user history. Furthermore, to mimic realistic online personalization settings, we restrict user-level personalization to a single training epoch.
% Per-Pcs and PriME involve three and two task-adaptive training stages, respectively (excluding the personalization stage), each of which consumes task-adaptive samples. In contrast, our method does not rely on task-adaptive training data and uses only target-user histories during personalization.

% \begin{table}[t]
% \centering
% \small
% \resizebox{\columnwidth}{!}{
% \begin{tabular}{lcc}
% \toprule
% Stage & Avg. Reduction (\%) & Max Reduction (\%) \\
% \midrule
% Task-adaptive training data & 99.3 & 99.7 \\
% Personalization user history & 89.5 & 97.2 \\
% \bottomrule
% \end{tabular}
% }
% \caption{Average and maximum data reduction achieved by our method across all LaMP tasks.}
% \label{tab:data_reduction_summary}
% \end{table}

\begin{table}[t]
\centering
\caption{Few-shot scenario statistics for the LaMP benchmark.
The table compares the scale of task-level data and user-level history
between full-data and few-shot settings.}
\label{tab:fewshot-scenario}

\resizebox{\columnwidth}{!}{%
\begin{tabular}{l cc cc}
\toprule
\textbf{Task}
& \multicolumn{2}{c}{\textbf{Full-data}}
& \multicolumn{2}{c}{\textbf{Few-shot}} \\
\cmidrule(lr){2-3} \cmidrule(lr){4-5}

& \shortstack{\textbf{\# Task Data}}
& \shortstack{\textbf{\# Avg. User Data}}
& \shortstack{\textbf{50 Task Data (\%)}}
& \shortstack{\textbf{10 User Data (\%)} } \\
\midrule

LaMP-1 & 7,122  & 147.16 & 0.70 & 6.8  \\
LaMP-2 & 2,810  & 37.25  & 1.78 & 26.8 \\
LaMP-3 & 20,129 & 360.61 & 0.25 & 2.8  \\
LaMP-4 & 8,821  & 155.93 & 0.57 & 6.4  \\
LaMP-5 & 14,464 & 144.04 & 0.35 & 6.9  \\
LaMP-7 & 13,327 & 77.17  & 0.38 & 13.0 \\

\midrule
\textbf{Average}
& \textbf{11,112}
& \textbf{153.69}
& \textbf{0.67}
& \textbf{10.4} \\

\bottomrule
\end{tabular}%
}
\end{table}

\paragraph{Baselines.}
We compare PRISP against a non-personalized base LLM (NP), as well as personalized base LLMs using prompt-based retrieval (RAG)  \citep{Salemi2024LaMP} and profile-based augmentation (PAG)  \citep{Richardson2023Integrating}. 
%We additionally consider RAG + PAG, which incorporates both retrieved history items and profile summaries into the input prompt. 
For history retrieval, we use BM25 \citep{trotman2014improvements} to select relevant user history items, with the number of retrieved items set to one. Following prior work, we apply RAG and PAG to all subsequent baselines.

We further evaluate PEFT-based personalization methods, including OPPU \citep{tan2024democratizing}, which serves as an upper bound in personalization scenario, and Per-Pcs \citep{tan2024personalized}, which constructs personalized modules by merging LoRA adapters from sharer users. 
We also include PriME \citep{kim2025personalized}, a gradient-free evolutionary framework that personalizes LLMs by improving task performance while explicitly controlling privacy leakage.

 We largely follow the hyperparameter settings of Per-Pcs, and PriME, except for the modifications explicitly described in Appendix~\ref{sec:appendix-exp_details}.

\begin{table*}[t]
\centering
\caption{Main results on the LaMP benchmark under a few-shot personalization setting.
% All baselines require access to task-specific data during the personalization process, whereas \textbf{Ours operates without task data}.
All baselines require access to task-specific data during the personalization process, whereas \textbf{PRISP (Ours) operates without task-level training data}, relying solely on the target user's own history for personalization.
When computing \textbf{Avg.}, we first aggregate metrics within each task as follows:
$(\mathrm{Acc}+\mathrm{F1})/2$ for LaMP-1/2,
$1-(\mathrm{MAE}+\mathrm{RMSE})/2$ for LaMP-3,
and $(\mathrm{R\text{-}1}+\mathrm{R\text{-}L})/2$ for LaMP-4/5/7,
and then average the resulting task-level scores. 
% Avg. is reported as a coarse summary across heterogeneous tasks. 
Arrows indicate whether higher ($\uparrow$) or lower ($\downarrow$) values are better.}
\resizebox{\textwidth}{!}{
\begin{tabular}{l c c|cccccccccccc|c}
\toprule
\multirow{3}{*}{\textbf{Method}}
& \multirow{3}{*}{\shortstack{\textbf{Shared}\\\textbf{Params.}}}
& \multirow{3}{*}{\shortstack{\textbf{Requires}\\\textbf{Task Data}}}
& \multicolumn{2}{c}{\makecell[c]{\textbf{LaMP-1}: Personal.\\Citation Identification}}
& \multicolumn{2}{c}{\makecell[c]{\textbf{LaMP-2}: Personal.\\Movie Tagging}}
& \multicolumn{2}{c}{\makecell[c]{\textbf{LaMP-3}: Personal.\\Product Rating}}
& \multicolumn{2}{c}{\makecell[c]{\textbf{LaMP-4}: Personal.\\News Headline Gen.}}
& \multicolumn{2}{c}{\makecell[c]{\textbf{LaMP-5}: Personal.\\Scholarly Title Gen.}}
& \multicolumn{2}{c}{\makecell[c]{\textbf{LaMP-7}: Personal.\\Tweet Paraphrasing}}
&  \\   % ← Avg 자리 비움
\cmidrule(lr){4-5}
\cmidrule(lr){6-7}
\cmidrule(lr){8-9}
\cmidrule(lr){10-11}
\cmidrule(lr){12-13}
\cmidrule(lr){14-15}
&
&
&
Acc $\uparrow$ & F1 $\uparrow$
& Acc $\uparrow$ & F1 $\uparrow$
& MAE $\downarrow$ & RMSE $\downarrow$
& R-1 $\uparrow$ & R-L $\uparrow$
& R-1 $\uparrow$ & R-L $\uparrow$
& R-1 $\uparrow$ & R-L $\uparrow$
& \textbf{Avg. $\uparrow$} \\
\midrule

NP        & $\times$ & Yes & 0.448 & 0.409 & 0.084 & 0.049 & 0.464 & 0.844 & 0.132 & 0.114 & 0.435 & 0.361 & 0.441 & 0.395 & 0.297 \\
RAG       & $\times$ & Yes & 0.456 & 0.337 & 0.333 & 0.224 & 0.430 & 0.841 & 0.129 & 0.113 & 0.434 & 0.374 & \textbf{0.447} & \textbf{0.397} & 0.331 \\
PAG       & $\times$ & Yes & 0.488 & 0.404 & 0.284 & 0.172 & 0.577 & 1.040 & 0.135 & 0.116 & 0.298 & 0.248 & 0.381 & 0.346 & 0.271 \\
RAG + PAG      & $\times$ & Yes & 0.456 & 0.337 & 0.369 & 0.274 & 0.446 & 0.841 & \textbf{0.138} & 0.120 & 0.437 & 0.377 & 0.399 & 0.362 & 0.332 \\
Per-Pcs   & $\checkmark$ & Yes & 0.480 & 0.421 & 0.362 & 0.240 & 0.420 & 0.819 & 0.134 & 0.116 & 0.420 & 0.357 & 0.222 & 0.208 & 0.310 \\
PriME     & $\checkmark$ & Yes & 0.504 & 0.414 & 0.362 & 0.240 & 0.430 & 0.827 & 0.135 & 0.116 & 0.410 & 0.340 & 0.325 & 0.288 & 0.323 \\
OPPU      & $\times$ & Yes & 0.472 & 0.464 & 0.373 & 0.245 & 0.410 & 0.791 & \textbf{0.138} & 0.121 & \textbf{0.441} & \textbf{0.379} & 0.437 & 0.391 & 0.355 \\
\midrule
\textbf{PRISP (Ours)}
          & $\times$ & \textbf{No}
          & \textbf{0.520} & \textbf{0.474}
          & \textbf{0.428} & \textbf{0.305}
          & \textbf{0.339} & \textbf{0.697}
          & \textbf{0.138} & \textbf{0.122}
          & 0.417 & 0.363 
          & 0.420 & 0.382
          & \textbf{0.378} \\
\bottomrule
\end{tabular}
\textbf{}}
\label{tab:main-result-qwen-only}
\end{table*}

\paragraph{Evaluation.}
For evaluation, we employ task-specific metrics consistent with the LaMP benchmark \citep{Salemi2024LaMP}. Specifically, we report Accuracy and F1 score for text classification tasks (LaMP-1 and LaMP-2), mean absolute error (MAE) and root mean squared error (RMSE) for the ordinal classification task (LaMP-3), and ROUGE-1 (R-1) and ROUGE-L (R-L) for text generation tasks (LaMP-4/5/7). Lower values indicate better performance for RMSE and MAE in LaMP-3, whereas higher values correspond to better performance for all other evaluation metrics.

\subsection{Results}
\paragraph{Overall Performance.}

Table~\ref{tab:main-result-qwen-only} presents the main results on the LaMP benchmark under a few-shot setting.
A key distinction is that all baselines require task-specific training data, whereas PRISP operates without accessing \textit{any} task data.

Across tasks, PRISP achieves the highest average score among all compared approaches. On classification tasks (LaMP-1 and LaMP-2), PRISP achieves the highest accuracy and F1 scores. For personalized product rating (LaMP-3), PRISP substantially reduces both MAE and RMSE, resulting in the best score among all methods. On text generation tasks (LaMP-4/5/7), PRISP achieves solid performance across all benchmarks, remaining competitive with strong baselines.

Overall, these results highlight that PRISP not only removes the dependency on task data, but also delivers strong and well-balanced performance across diverse personalization tasks, underscoring its effectiveness and practicality in data-restricted personalization scenarios.

\begin{figure}
    \centering
    \includegraphics[width=\columnwidth]{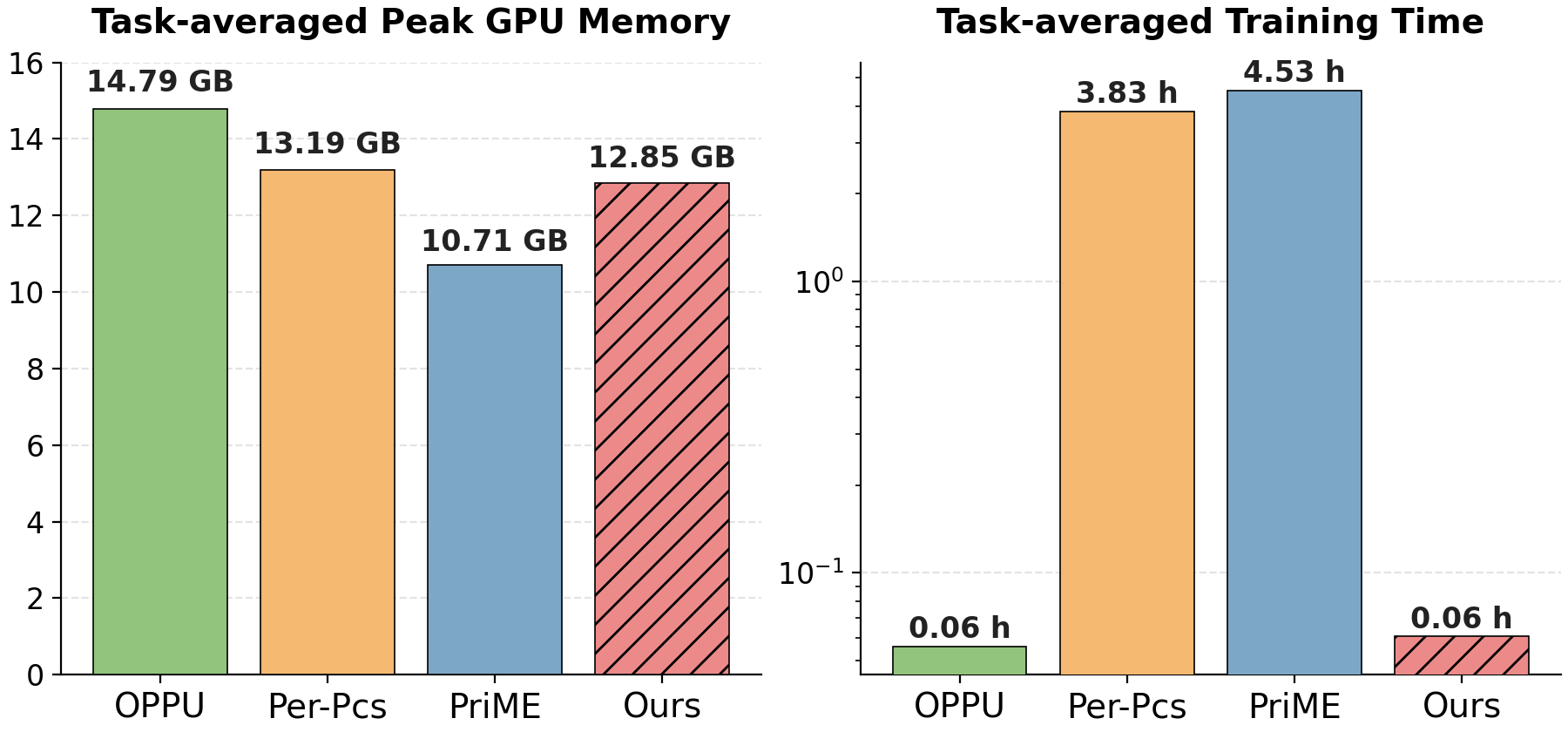}
    \caption{Task-averaged peak GPU memory usage and training time on the LaMP benchmark, where values are averaged across all tasks. PRISP (Ours) achieves competitive memory consumption while reducing training time by orders of magnitude compared to prior personalization approaches.}

    \label{fig:cost_plot}
\end{figure}

\paragraph{Computational Efficiency.}

Figure~\ref{fig:cost_plot} summarizes the computational costs of different personalization methods in terms of average peak GPU memory usage and training time. As shown in the figure, Per-Pcs and PriME incur substantially longer training time compared to other methods, which is attributed to their reliance on optimizing multiple sharer LoRAs during personalization.

In Per-Pcs, personalization requires selecting and composing a subset of sharer LoRAs based on user–sharer similarity, while PriME further introduces additional overhead through greedy evolutionary optimization that evaluates a large number of candidate LoRA modules via repeated inference. 

In contrast, OPPU and our PRISP dramatically reduce training time, achieving approximately two orders-of-magnitude speedups over Per-Pcs and PriME, by eliminating the need of multiple sharer LoRAs and instead adopting a direct and lightweight adaptation strategy. Furthermore, compared to OPPU, PRISP improves GPU memory efficiency  during personalization by freezing the input LoRA. 

\begin{figure}
    \centering
    \includegraphics[width=0.75\columnwidth]{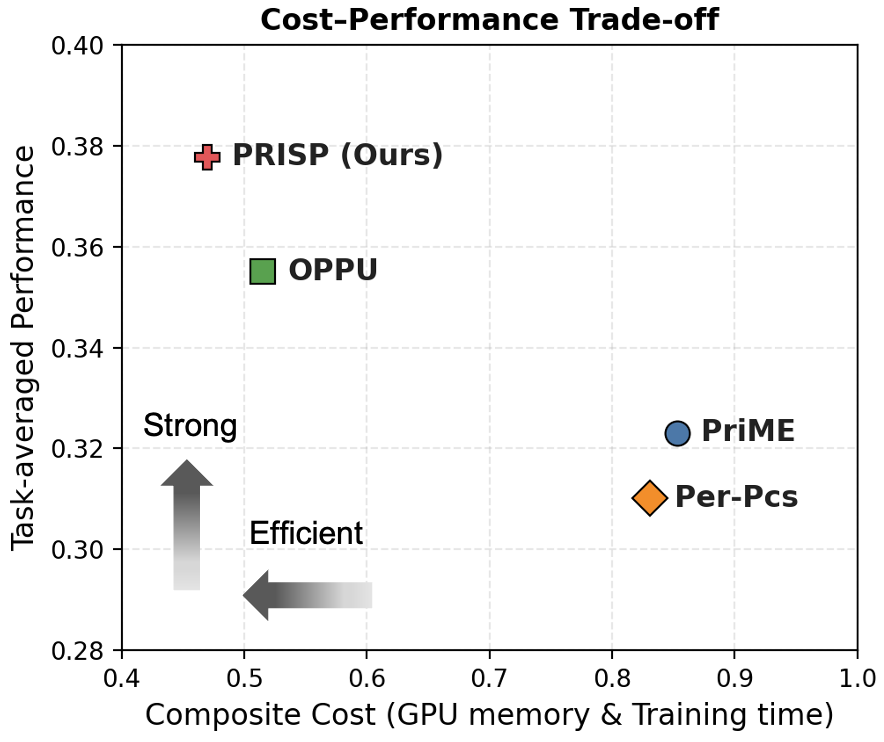}
    \caption{Cost-performance trade-off on the LaMP benchmark under a few-shot setting. Composite cost is computed by max-normalizing GPU memory usage and training time, and averaging them with equal weights.}

    \label{fig:cost-perform-trade-off}
\end{figure}

Regarding the hypernetwork overhead, task-aware LoRA parameters are generated with a single inference pass of the hypernetwork, which incurs negligible latency and does not increase peak GPU memory usage. Further details are provided in Appendix~\ref{sec:appendix-exp_details}. Overall, our approach achieves a more favorable cost–performance trade-off, as illustrated in Figure~\ref{fig:cost-perform-trade-off}.

\paragraph{Privacy.}
Table~\ref{tab:main-result-qwen-only} reports whether each method requires access to parameters shared across users, as indicated in the \textit{Shared Params} column. 
Per-Pcs and PriME rely on LoRA parameters trained on other users (\textit{sharers}) who are considered similar to the target user. This design assumes that user-specific LoRA modules are accessible with their consent. However, such parameters encode user information and therefore pose a risk of privacy leakage \citep{carlini2019secret,carlini2021extracting}. In contrast, PRISP does not leverage sharer LoRA modules and thus avoids cross-user parameter sharing. 

Moreover, unlike all baselines, PRISP does not require task data. In the LaMP benchmark, task data is constructed from other users’ data, meaning that its use implicitly introduces additional privacy concerns. Similar issues arise in real-world settings, where task data typically comes from external or third-party user sources. By operating without task data, PRISP removes this source of privacy risk altogether. As a result, our approach is privacy issue-free by design.

\begin{table}[t]
\centering
\caption{Results on LaMP under the standard full-data personalization setting.
For each task, scores are obtained by averaging the two corresponding evaluation metrics.}
\resizebox{\columnwidth}{!}{
\begin{tabular}{l c c c c c c | c}
\toprule
\textbf{Method}
& \textbf{LaMP-1} $\uparrow$
& \textbf{LaMP-2} $\uparrow$
& \textbf{LaMP-3} $\downarrow$
& \textbf{LaMP-4} $\uparrow$
& \textbf{LaMP-5} $\uparrow$
& \textbf{LaMP-7} $\uparrow$
& \textbf{Avg.} $\uparrow$ \\
\midrule
NP
& 0.599 & 0.238 & 0.510 & 0.133 & 0.384 & 0.431
& 0.379 \\
RAG
& 0.584 & 0.390 & 0.436 & 0.143 & 0.409 & 0.449
& 0.423 \\
PAG
& 0.702 & 0.389 & 0.482 & 0.139 & 0.392 & 0.433
& 0.429 \\
RAG + PAG
& 0.695 & 0.445 & 0.444 & 0.147 & 0.411 & 0.469
& 0.454 \\

Per-Pcs
& 0.698 & 0.454 & 0.483 & 0.139 & 0.397 & 0.444
& 0.442 \\
PriME
& 0.727 & 0.445 & 0.476 & 0.143 & 0.396 & 0.447
& 0.447 \\
OPPU
& \textbf{0.735} & 0.461 & 0.433 & \textbf{0.150} & \textbf{0.414} & 0.455
& 0.464 \\
\midrule
\textbf{PRISP (Ours)}
& 0.704 & \textbf{0.529} & \textbf{0.424} & 0.147 & 0.408 & \textbf{0.470}
& \textbf{0.472} \\
\bottomrule
\end{tabular}
}
\label{tab:lamp_full_singlecol_avg}
\end{table}

\paragraph{Full-Data Personalization.}
In addition to the few-shot personalization setting, we also consider the full-data setting, which corresponds to the standard evaluation scenario commonly adopted in prior personalization studies. In this setting, all available training data in the LaMP benchmark are utilized, following the data usage protocol of each respective method. For our approach, the task data are used to further train the anchor LoRA generated by the hypernetwork.

As shown in Table~\ref{tab:lamp_full_singlecol_avg}, OPPU exhibits strong overall performance across tasks, reflecting its ability to leverage task- and user-specific LoRA modules trained via gradient-based optimization with access to the full training data. Since OPPU introduces separate LoRA modules for each task and each user and optimizes them using all available data, it is commonly regarded as an upper bound in personalization settings.

Despite having substantially fewer trainable parameters due to freezing, PRISP achieves competitive performance across all tasks and attains the best average score among the compared approaches. These results demonstrate that PRISP remains effective not only in few-shot settings but also in the standard full-data personalization scenario.

\subsection{Ablation Studies}
\label{sec:ablation studies}

For all ablation studies and analysis, we report results obtained without RAG \citep{Salemi2024LaMP} or PAG \citep{Richardson2023Integrating} retrieval methods, in order to isolate the core effects of each method.

\paragraph{Task-Aware Anchor LoRA.}
\begin{figure}
    \centering
    \includegraphics[width=\columnwidth]{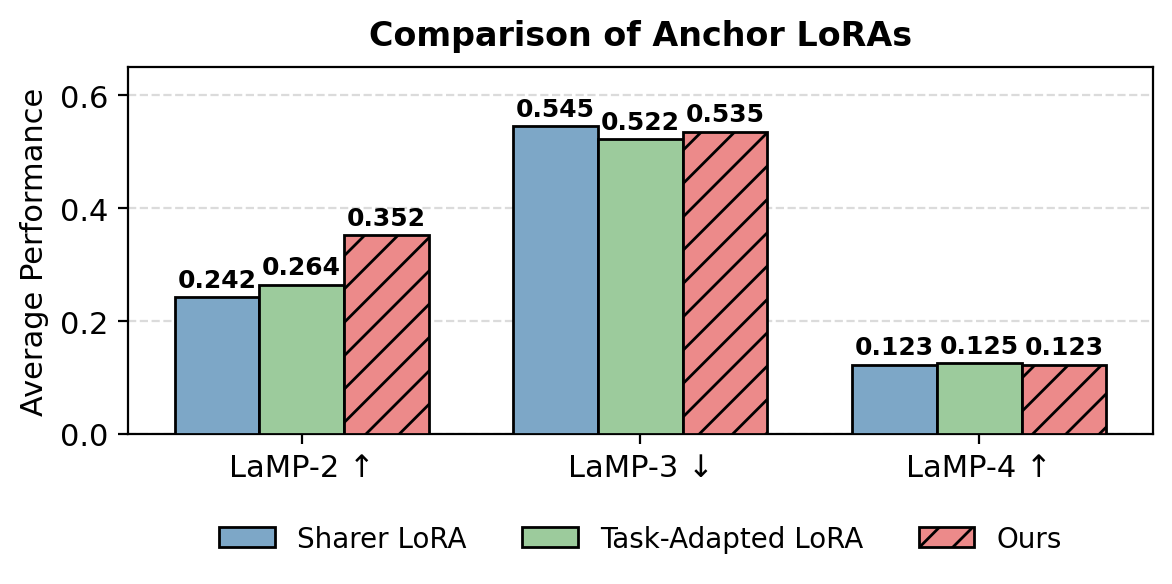}
    \caption{Comparison of anchor LoRAs used for personalization.
We compare three anchor LoRAs: (i) a sharer LoRA selected most frequently during PriME optimization, (ii) a task-adapted LoRA trained on 50 task-specific samples, and (iii) our task-aware anchor LoRA generated via a hypernetwork. The reported values denote average performance computed by aggregating the task-specific evaluation metrics for each task.}

    \label{fig:Effects of our Anchor LoRA}
\end{figure}
In this section, we analyze whether a task-aware anchor LoRA generated via a hypernetwork can serve as a strong initialization for personalization. We compare our anchor LoRA against two baselines: (i) a representative sharer LoRA from PriME, selected as the most frequently chosen sharer across target users, and (ii) a task-adapted LoRA trained on 50 task-specific samples.

As shown in Figure~\ref{fig:Effects of our Anchor LoRA}, our anchor LoRA achieves competitive performance with both the sharer LoRA and the task-adapted LoRA, despite not relying on user-embedding similarity or task data. This indicates that a task-aware anchor LoRA alone provides a reliable initialization for personalization.

Moreover, we observe that sharer LoRAs do not consistently outperform other anchors and can even underperform in certain tasks, suggesting limited generalization caused by user-specific bias, which is a critical limitation in realistic personalization scenarios where user preferences are highly diverse. 

% Overall, these results demonstrate that a general task-aware anchor LoRA offers a robust, privacy-friendly, and efficient alternative to user-specific sharer LoRAs or task-adapted LoRAs for personalization.

\begin{table}[t]
\centering
\caption{Comparison with alternative LoRA initialization strategies under the few-shot personalization setting.}
\label{tab:init_comparison}
\small
\resizebox{\columnwidth}{!}{
\begin{tabular}{lcccccc}
\toprule
\multirow{2}{*}{Method} & \multicolumn{2}{c}{LaMP-2} & \multicolumn{2}{c}{LaMP-3} & \multicolumn{2}{c}{LaMP-7} \\
\cmidrule(lr){2-3} \cmidrule(lr){4-5} \cmidrule(lr){6-7}
& Acc $\uparrow$ & F1 $\uparrow$ & MAE $\downarrow$ & RMSE $\downarrow$ & R-1 $\uparrow$ & R-L $\uparrow$ \\
\midrule
PiSSA & 0.379 & 0.250 & 0.404 & 0.770 & 0.339 & 0.301 \\
LoRA-GA & 0.340 & 0.211 & 0.510 & 0.939 & 0.318 & 0.293 \\
\textbf{PRISP (Ours)} & \textbf{0.428} & \textbf{0.305} & \textbf{0.339} & \textbf{0.697} & \textbf{0.420} & \textbf{0.382} \\
\bottomrule
\end{tabular}
}
\end{table}

\paragraph{Alternative Initialization Strategies.} We compare PRISP against two SVD-based LoRA initialization methods: PiSSA~\citep{meng2024pissa}, which initializes LoRA matrices from the principal singular components of the pre-trained weights, and LoRA-GA~\citep{wang2024loraga}, which constructs a gradient-aligned low-rank subspace using user-specific calibration gradients. As shown in Table~\ref{tab:init_comparison}, both methods underperform compared to PRISP across all evaluated tasks in the few-shot regime. We attribute this to the instability of SVD-based approaches when operating on extremely limited data, whereas PRISP conditions initialization on semantic task information via the hypernetwork, providing a more robust inductive bias independent of user data quantity.

\paragraph{Bridge and Freezing.}
We analyze different personalization strategies by comparing several fine-tuning variants that differ in which components are trained, as summarized in Table \ref{tab:Comparison of bridge}.

We observe that freezing the input LoRA (No Bridge) achieves performance comparable to training both LoRA matrices (Full LoRA). This suggests that in extreme few-shot settings, updating fewer parameters is sufficient. Comparing the No Bridge variant with PRISP illustrates the role of the bridge matrix, where introducing a small trainable bridge leads to improved personalization performance over training the output LoRA alone. Training only the bridge matrix (Bridge Only) mostly underperforms other variants, indicating that the bridge alone lacks sufficient expressive capacity. 

% and must be jointly trained with at least one LoRA component. Overall, these results validate our design choice of freezing the input LoRA while jointly training the output LoRA and bridge matrix for effective and efficient few-shot personalization.

\begin{table}[t]
\caption{Comparison of personalization strategies on LaMP benchmarks.
We compare fine-tuning variants that differ in which components of the LoRA-based update are trained.
\textit{Full LoRA} trains the standard low-rank update $BA$.
\textit{No Bridge} freezes $A$ and trains only $B$.
\textit{Bridge Only} trains the bridge-augmented update $BCA$ with $BA$ frozen.
\textit{PRISP} trains $BC$ while freezing $A$ in the bridge-augmented formulation $BCA$.}

\centering
\footnotesize
\setlength{\tabcolsep}{4pt}
\begin{tabular}{l l
              c c c c}
\toprule
Task & Metric & {Full} & {No} & {Bridge} & {\textbf{PRISP}} \\
     &        & {LoRA} & {Bridge} & {Only} &  {\textbf{(Ours)}}    \\
\midrule
\multirow{2}{*}{LaMP-2}
 & Acc $\uparrow$   & 0.307 & 0.267 & 0.209  & \textbf{0.412} \\
 & F1 $\uparrow$    & 0.221 & 0.198 & 0.083  & \textbf{0.292} \\
\midrule
\multirow{2}{*}{LaMP-3}
 & MAE $\downarrow$   & 0.378 & 0.379 & 0.944 & \textbf{0.358} \\
 & RMSE $\downarrow$    & 0.715 & \textbf{0.705} & 1.402 & 0.713 \\
\midrule
\multirow{2}{*}{LaMP-4}
 & R-1 $\uparrow$   & \textbf{0.136} & 0.132 & 0.103 & 0.131 \\
 & R-L $\uparrow$    & \textbf{0.117} & 0.114 & 0.092 & 0.115 \\
\bottomrule
\label{tab:Comparison of bridge}
\end{tabular}

\end{table}

\subsection{Analysis}
\label{sec:Analysis}

\paragraph{Data Size Analysis.}
We evaluate how effectively each method adapts to limited user-specific data by varying the amount of target user data from one shot to 50 shots under a few-shot regime.

\begin{figure}
    \centering
    \includegraphics[width=\columnwidth]{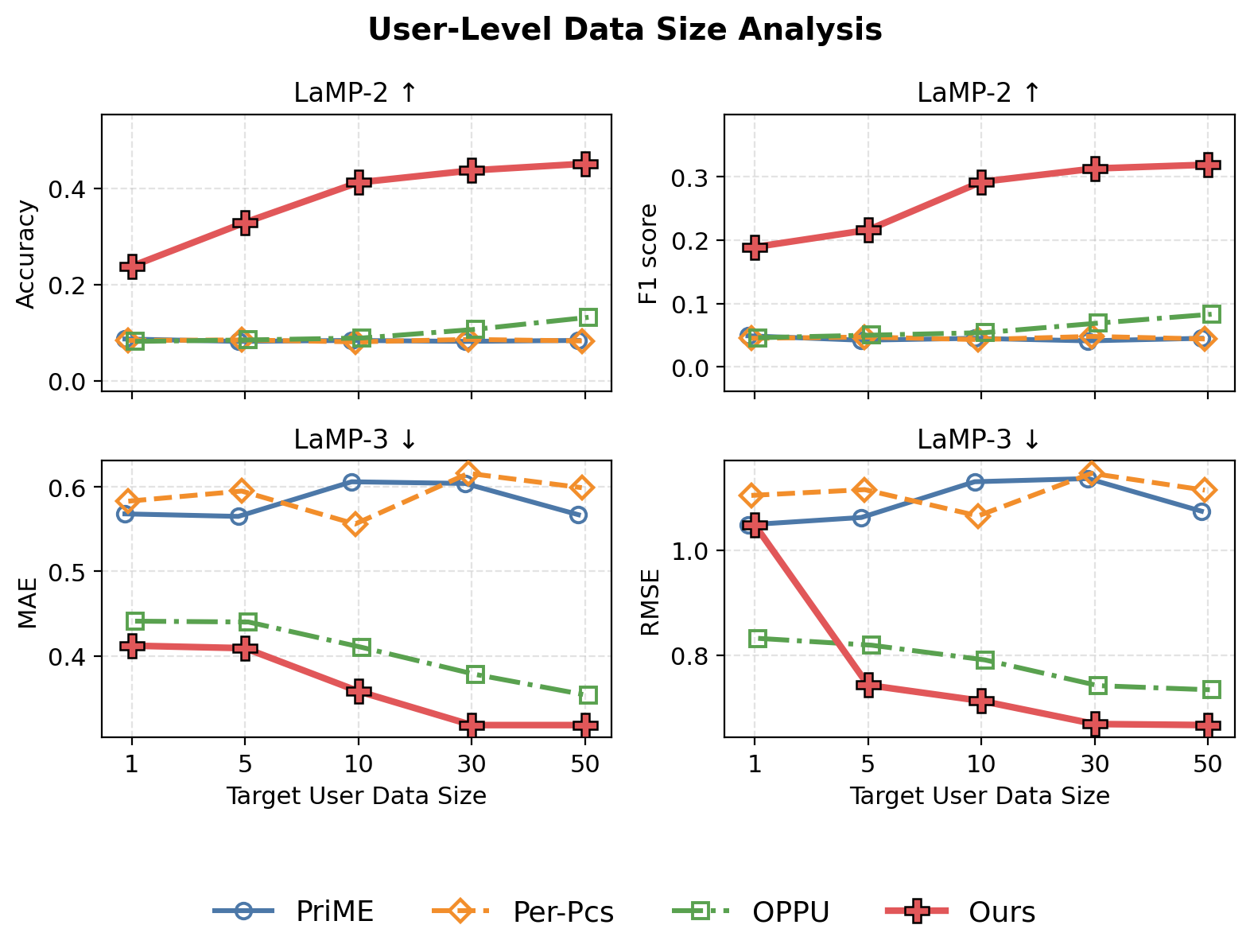}
    \caption{Performance comparison under varying target user data sizes (1, 5, 10, 30, and 50).  For baseline methods, the task-level data budget is fixed to 50 samples..}

    \label{fig:datasize}
\end{figure}

In few-shot settings, PRISP operates with minimal supervision, relying solely on the available target user data for personalization. Despite this constraint, PRISP demonstrates strong performance even with a small number of target user samples, as shown in Figure \ref{fig:datasize}. Moreover, PRISP exhibits a markedly steeper performance slope as the number of target user samples increases, indicating fast adaptation in few-shot personalization scenarios. In contrast, competing methods show little to no performance improvement as additional user data is provided, suggesting slower adaptation and limited practicality in realistic settings where only a small amount of user-specific data is available.

% Overall, our method demonstrates strong sample efficiency, achieving competitive or superior performance in few-shot regimes and adapting more rapidly as additional user data becomes available.

\paragraph{Generalization to Unseen Tasks.}

In realistic personalization scenarios, models are often required to generalize to \emph{arbitrary} or previously unseen personalization tasks. While existing benchmarks typically assume that sufficient task-level data and access to task-adapted sharer LoRAs are available for each target task, this assumption is not always practical. 
% For example, a model may be trained and deployed on a single task (e.g., LaMP-1) and later be required to adapt to a new task (e.g., LaMP-2) without access to task-adapted sharer LoRAs or extensive task-level supervision.

In this section, we demonstrate that, unlike PRISP, prior approaches that rely on sharer LoRAs exhibit limited generalization to unseen personalization tasks. To analyze this limitation, we consider a cross-task personalization setting in which sharer LoRAs trained on an initial task (referred to as the \emph{source task}) are directly reused and applied to different target tasks.

We define \textit{adaptability} as the relative performance retained when adapting personalization from a source task to an unseen target task. Formally, adaptability is computed as
\begin{equation}
\text{Adaptability} =
\frac{
\mathrm{Perf}\left(
\mathcal{S}_{\text{source}} \rightarrow \text{Target Task}
\right)
}{
\mathrm{Perf}\left(
\mathcal{S}_{\text{target}} \rightarrow \text{Target Task}
\right)
}
\label{eq:adaptability}
\notag
\end{equation}
where $\mathcal{S}_{\text{source}}$ and $\mathcal{S}_{\text{target}}$ denote sharer LoRA modules trained on the source and target tasks, respectively. 
$\mathrm{Perf}(\mathcal{S} \!\rightarrow\! \text{Target Task})$ measures the performance on the target task when using sharer LoRA $\mathcal{S}$, averaged over task-specific evaluation metrics.
Intuitively, adaptability quantifies how well a sharer LoRA trained on a different task can be reused for personalization on an unseen target task.

\begin{figure}
    \centering
    \includegraphics[width=\columnwidth]{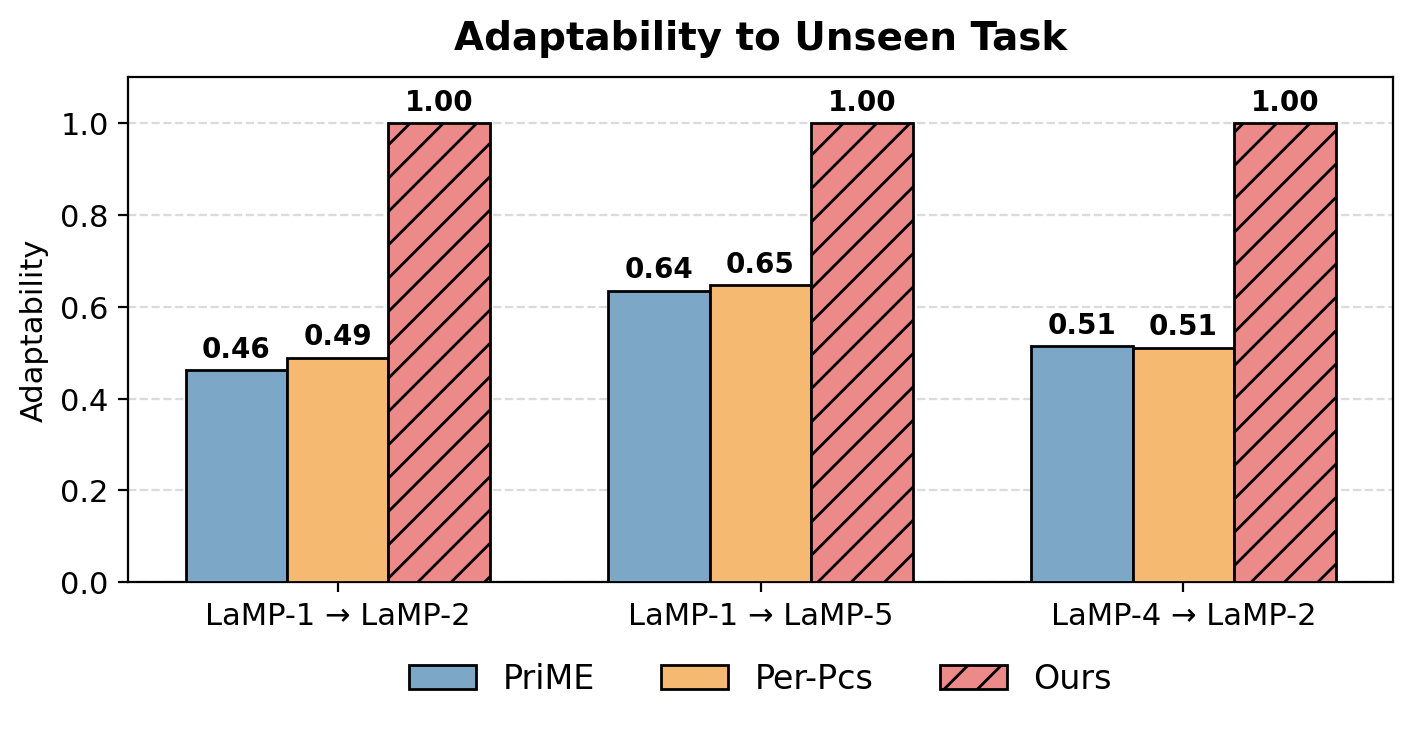}
    \caption{Adaptation to unseen personalization tasks. Columns represent source-to-target task adaptation scenarios, where sharer LoRAs trained on the source task are directly reused without additional retraining, followed by adaptation using only the target user data.}

    \label{fig:Transfer across unseen task}
\end{figure}

As shown in Figure~\ref{fig:Transfer across unseen task}, sharer-LoRA-based methods exhibit low adaptability, with values around 50--60\%. This degradation suggests that personalization knowledge learned in a source task does not effectively generalize across tasks and that target-task sharer LoRAs are required to maintain performance in these approaches. In contrast, PRISP is regarded as having an adaptability of 100\%, because it can generate task-aware LoRAs in a zero-shot manner from task descriptions via a hypernetwork, enabling direct adaptation to unseen target tasks without relying on task data or task-specific sharer LoRAs. 

% Overall, these results highlight that while sharer LoRA-based methods exhibit poor task generalization, as reflected by their low adaptability, our approach enables robust personalization in settings where adaptation to arbitrary new tasks is required, without requiring task-specific sharer LoRAs.

\begin{figure}[t]
    \centering
    \includegraphics[width=0.95\linewidth]{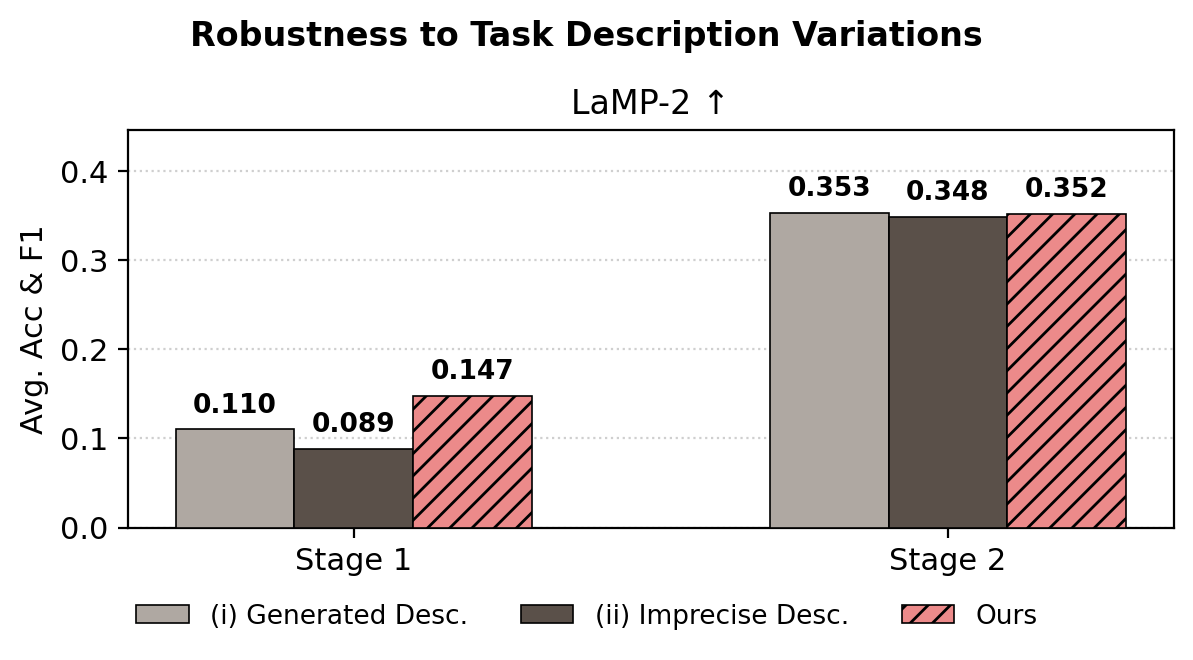}
    \caption{
   Comparison of task description variations in LaMP-2.
Stage 1 evaluates task-aware LoRA adapters generated from different task descriptions without personalization, while Stage 2 (personalization) continues from the corresponding Stage~1 adapters and applies personalization. Performance is averaged using task-specific metrics.
    }
    \label{fig:task_robustness}
\end{figure}

\paragraph{Robustness to Task Description Variations.}
\label{Ablation:robustness to task description variations}
In Figure~\ref{fig:task_robustness}, we evaluate robustness by comparing two alternative task descriptions with our reference descriptions. 
We consider two variants: (i) task descriptions generated using the Text-to-LoRA prompt~\citep{charakorn2025texttolora}, which preserve the semantic meaning of the reference description but differ stylistically; and (ii) task descriptions selected based on the highest cosine similarity to the reference descriptions from the Text-to-LoRA hypernetwork training set, simulating slightly imprecise user-provided task descriptions. 
Exact descriptions are provided in Appendix~\ref{sec:appendix-exp_details}.
% (iii) task descriptions from different LaMP tasks that are irrelevant to the target task
% whereas the third variant is intentionally irrelevant to the target task

For each task description variant, we analyze performance in two sequential stages, corresponding to the two columns in Figure~\ref{fig:task_robustness}. In Stage~1, we apply only the task-aware LoRA generated from each description to the base model and evaluate performance without personalization. In Stage~2, we continue from the corresponding task-aware LoRA and perform personalization, allowing the model to adapt using target user data.

In Stage~1, performance varies across task description variants, with the lowest performance observed for the slightly imprecise task description. In contrast, in Stage~2, performance across all variants consistently converges to a similar level, suggesting that PRISP mitigates task description variations and is robust to such variations.

% \subsubsection{Bridge Analysis}
% How different between W/o Bridge, BA, and ours.
% 1. Bridge value after train.
% 2. Gradient norm during training step of Bridge and B vs that of B (W/o Bridge)

% => Bridge or B had large grad norm in early training step than  B(w/o bridge) and this may affect the fast adaptation.

% What is the difference between B,P matrix combined vs only B matrix(P merged)

% \begin{figure}
%     \centering
%     \includegraphics[width=\columnwidth]{5.4.4.png}
%     \caption{
% Robustness to task description variations across different adaptation stages.
% Each subplot reports performance under three stages; 
% \textit{Base Model}: the unadapted backbone without LoRA, 
% \textit{Inference}: direct inference after merging the LoRA generated by Text-to-LoRA, and 
% \textit{Personalization}: performance after applying user-specific personalization.
% We compare four task description variants: the reference LaMP prompt, a Text-to-LoRA-generated prompt, the most similar prompt selected by cosine similarity, and an unrelated prompt from a different LaMP task.
% }

%     \label{fig:placeholder4}
% \end{figure}

% \subsubsection{(Option)Merging based method(Per-PCS, PriME) vs Training based method(OPPU, Ours)}
% K=0,1,2,4, PAG Comaprison

% Ours, OPPU: not that affected by k=1,2,4 because the samples are directly trained on the paramter
% PriME, Per-PCS: same???

% (Option)
% Why does general lora perform better in this setting? The distance is smaller so that they can be merged safely but freely with domix than just interpolating. ...

\section{Conclusion}
This paper addresses practical LLM personalization under limited user data, constrained computation, and strict privacy requirements. We propose PRISP, a lightweight and privacy-safe framework that operates without task data by generating task-aware LoRA modules from task descriptions via a hypernetwork and adapting to users with lightweight modules. Experiments demonstrate that PRISP achieves strong overall performance compared to prior methods, while reducing computational overhead and avoiding privacy risks. Through extensive analysis, we further demonstrate that PRISP remains effective across diverse few-shot regimes, generalizes well to unseen tasks, and is robust to variations in task descriptions. These results indicate that PRISP is a broadly applicable framework suitable for realistic personalization scenarios.

\section*{Limitations}
Although the proposed method is designed under a realistic personalization setting, our current evaluation focuses on static personalization scenarios and does not explicitly consider continual settings where user data and preferences evolve dynamically over time. In addition, our framework assumes the availability of a hypernetwork capable of generating meaningful task-aware anchor LoRA parameters from task descriptions. While this design enables task-data-free personalization, the quality of personalization may depend on the expressiveness of the hypernetwork. Extending our framework to continual personalization settings and further improving the expressiveness of the hypernetwork, for example through additional training, are promising directions for future work.

\section*{Ethical Considerations}

Our work involves personalization using user-specific data, which may contain sensitive information. To mitigate privacy risks, PRISP is designed to operate under strict data minimization principles, avoiding parameter sharing across users and eliminating the need to access other users’ data during personalization. All experiments are conducted on benchmark datasets with anonymized user identifiers, and no real-world personally identifiable information is collected or released. In addition, an AI assistant (ChatGPT) was used for minor grammar and writing refinement.

\section*{Acknowledgments}

This work was supported in part by National Research Foundation of Korea (NRF) grant [No. RS-2025-02263628, No. RS-202300265406], the Institute of Information \& communications Technology Planning \& Evaluation (IITP) grants [RS-2021-II212068,
RS-2022-II220113, RS-2022-II220959, RS-2021-II211343], and the BK21 FOUR Education and
Research Program for Future ICT Pioneers (Seoul
National University), funded by the Korean government (MSIT). It was also supported by Mobile
eXperience (MX) Business, Samsung Electronics
Co., Ltd and AOARD Grant No. FA2386-23-1-4079.

% \section*{Acknowledgments}

\bibliography{custom}    % References.bib 파일을 사용

\appendix

\section{Datasets}
\label{sec:appendix-datasets}
\subsection{Benchmark \& Task Details}
In this section, we give brief explanation of individual tasks in LaMP benchmark. LaMP-6 : Email subject generation task is excluded since it involves private data inaccessible.

\vspace{0.3em}
\paragraph{LaMP-1: Personalized Citation Identification.}
This is a binary text classification task which assesses the ability of a language model to determine the user preferred citation between two candidate papers, given user's history of past publications. User's profile contains title, abstract, and citation information of previous publications.

\paragraph{LaMP-2: Personalized Movie Tagging.}

This task is a personalized multi-label classification problem, where the model predicts tags a user would assign to a movie.
Given a movie description, the model outputs a tag out of 15 possible categories, reflecting the user’s subjective interpretation and tagging behavior.
The user profile contains previously tagged movies, along with the tags assigned by the user.

\paragraph{LaMP-3: Personalized Product Rating Prediction.}

This task focuses on personalized ordinal classification by predicting a user-specific rating for a given product review.
The model estimates the rating from 1 to 5 the target user would assign, reflecting subjective preferences rather than objective quality.
The user profile contains historical product reviews written by the user, paired with their corresponding ratings.

\paragraph{LaMP-4: Personalized News Headline Generation.}

This task evaluates personalized text generation by requiring the model to produce a news headline tailored to the target user’s writing style and topical emphasis.
Given the body of a news article, the model generates a headline that reflects both the article content and how the user typically frames similar news.
The user profile consists of previously written or read news articles paired with their corresponding headlines.

\paragraph{LaMP-5: Personalized Scholarly Title Generation.}

This is a personalized text generation task that requires generating an academic paper title aligned with the user’s writing style.
Given an abstract, the model generates a suitable title that reflects both the content of the paper and the user’s historical preferences.
The user profile consists of previously authored paper titles and abstracts.

\begin{table*}[t]
\caption{Dataset statistics across different data partitions used in our experiments.}

\centering
\small
\setlength{\tabcolsep}{6pt}
\renewcommand{\arraystretch}{1.15}
\begin{tabular}{lccc ccc ccc}
\toprule
\multirow{2}{*}{Task} &
\multicolumn{3}{c}{Task-adaptive data} &
\multicolumn{3}{c}{Sharer candidates} &
\multicolumn{3}{c}{Target users} \\
\cmidrule(lr){2-4}
\cmidrule(lr){5-7}
\cmidrule(lr){8-10}
& Users & \# Queries & Avg. Hist. &
Users & \# Queries & Avg. Hist. &
Users & \# Queries & Avg. Hist. \\
\midrule
LaMP-1 & 1472 & 1788 & 65.35  & 4316  & 5334  & 88.48  & 100 & 125  & 147.16 \\
LaMP-2 & 159  & 425  & 5.15   & 336   & 2385  & 12.28  & 100 & 2228 & 37.25  \\
LaMP-3 & 369  & 1235 & 14.62  & 1010  & 7586  & 31.32  & 100 & 3949 & 155.93 \\
LaMP-4 & 4500 & 5095 & 130.65 & 13400 & 15034 & 202.52 & 100 & 614  & 360.61 \\
LaMP-5 & 3303 & 3643 & 65.51  & 9811  & 10821 & 94.34  & 100 & 608  & 144.04 \\
LaMP-7 & 3023 & 3349 & 14.03  & 8971  & 9978  & 15.67  & 100 & 114  & 77.17  \\
\bottomrule
\end{tabular}

\label{tab:lamp_dataset_statistics}
\end{table*}

\paragraph{LaMP-7: Personalized Tweet Paraphrasing.}

This task examines personalized paraphrase generation by rewriting a tweet according to the user’s linguistic style and expression preferences.
Given an input tweet, the model generates a paraphrased version that preserves the original meaning while matching the user’s typical phrasing.
The user profile consists of historical tweet–paraphrase pairs produced by the user.

\subsection{Setup}
\paragraph{Data Splits.}
We adopt the refined dataset split proposed by Per-Pcs \citep{tan2024personalized}, which reorganizes the LaMP benchmark \citep{Salemi2024LaMP} into three disjoint subsets. Specifically, 25\% of users are used to construct the task-adaptive training data, 100 randomly sampled users are reserved for evaluation, and the remaining users are treated as sharer candidates whose data are used to train sharer LoRA modules. Table~\ref{tab:lamp_dataset_statistics} summarizes the resulting data statistics.

In the full-data setting, sharer candidate data are additionally included in the task-adaptive training set for methods that do not utilize sharer LoRAs, ensuring a fair comparison in terms of access to the total available data.

\paragraph{Prompt Details.}
The prompts used for the personalization tasks are provided in Table \ref{tab:personalization_prompts}. In addition, the prompts employed for generating user profiles are presented in Table \ref{tab: profile generation}.

\section{Experimental Details}
\label{sec:appendix-exp_details}
This section provides a detailed explanation of the
experiments. All experiments were conducted
using NVIDIA A100 GPUs with 80GB
of memory.

% 본문
\paragraph{Hyperparameters.}
Table \ref{tab:lamp_hparams} shows the hyperparameters used in our experiments. We adopt the hyperparameter settings reported in Per-Pcs for all baseline implementations whenever applicable. Due to the substantial training cost of PriME, we reduce its training budget to ensure a practical training time. We set the privacy constraint of PriME to 1.0 for all experiments, following the original paper, as it is reported to provide a favorable trade-off between performance and privacy preservation. For OPPU, we align the hyperparameters of the personalization stage with those used in PRISP.

\begin{table*}[t]
\caption{Hyperparameters for task adaptation, sharer-based methods Per-Pcs, PRiME, and \textbf{PRISP (Ours)} on LaMP.}

\centering
\small
\setlength{\tabcolsep}{4.2pt}
\renewcommand{\arraystretch}{1.25}
\begin{tabular}{
l
ccc | 
ccc | 
ccc | 
cc | 
cc | 
ccc
}
\toprule
\multirow{2}{*}{Task} &
\multicolumn{3}{c|}{Task Adaptation} &
\multicolumn{3}{c|}{Sharer PEFT} &
\multicolumn{3}{c|}{Sharer Gate} &
\multicolumn{2}{c|}{Per-Pcs} &
\multicolumn{2}{c|}{PRiME} &
\multicolumn{3}{c}{\textbf{PRISP (Ours)}} \\
\cmidrule(lr){2-4}
\cmidrule(lr){5-7}
\cmidrule(lr){8-10}
\cmidrule(lr){11-12}
\cmidrule(lr){13-14}
\cmidrule(lr){15-17}
& batch & ep & lr
& batch & ep & lr
& batch & step & lr
& top-$k$ & batch
& LoRA & budget
& batch & ep & lr \\
\midrule
LaMP-1 & 6 & 3 & 1e-4 & 16 & 1 & 1e-5 & 6 & 100 & 1e-5 & 1 & 16 & 5 & 10 & 16 & 1 & 2e-4 \\
LaMP-2 & 6 & 3 & 1e-4 & 6 & 3 & 2e-5 & 6 & 100 & 2e-5 & 3 & 16 & 3 & 5 & 16 & 1 & 1e-4 \\
LaMP-3 & 6 & 3 & 1e-4 & 2 & 2 & 1e-5 & 4 & 100 & 1e-5 & 1 & 6  & 5 & 10 & 6 & 1 & 1e-4 \\
LaMP-4 & 6 & 3 & 1e-4 & 10 & 3 & 2e-5 & 6 & 50  & 2e-5 & 1 & 16 & 3 & 5 & 16 & 1 & 5e-4 \\
LaMP-5 & 6 & 3 & 1e-4 & 3 & 2 & 2e-5 & 6 & 50  & 2e-5 & 1 & 10 & 3 & 10 & 10 & 1 & 1e-6 \\
LaMP-7 & 6 & 3 & 1e-4 & 16 & 2 & 1e-5 & 6 & 50  & 2e-5 & 2 & 16 & 3 & 5 & 16 & 1 & 3e-6 \\
\bottomrule
\end{tabular}

\label{tab:lamp_hparams}
\end{table*}

\paragraph{LoRA Configuration.}

We employ Low-Rank Adaptation (LoRA)~\cite{Hu2022LoRA} as a parameter-efficient fine-tuning method.
We set the rank of the low-rank decomposition to $r = 8$.
For PRISP and OPPU, we apply LoRA modules exclusively to the query and value projection layers (\texttt{q\_proj} and \texttt{v\_proj}) within the self-attention blocks,
whereas other baselines apply LoRA modules to the query, key, value, and output projection layers
(\texttt{q\_proj}, \texttt{k\_proj}, \texttt{v\_proj}, and \texttt{o\_proj}).
A dropout rate of 0.05 is applied to all LoRA layers, and bias terms in the LoRA parameters are disabled.

% All experiments are conducted under the causal language modeling objective.

% \paragraph{Retrieval and Profile Augmentation Settings.}
% Unless otherwise stated, all experiments are conducted without retrieval-augmented generation (RAG) or profile-augmented generation (PAG).
% In particular, all results and illustrations in Figure~1 are obtained under a retrieval- and profile-free setting to isolate the effect of parametric personalization.

\paragraph{Task Descriptions Used in Experiments.}

Table~\ref{tab:text_to_lora_task_descriptions} presents the task descriptions used in our main experiments.
These descriptions are adapted from the original LaMP prompt \citep{Salemi2024LaMP} with minor modifications.
The resulting task descriptions are provided as input to the hypernetwork to generate task-aware LoRA modules. Additional task descriptions used in Section \ref{Ablation:robustness to task description variations} for LaMP-2 is provided in Table \ref{tab:movie_tagging_task_description_variations}.

\paragraph{Detailed Results of the Main Experiments.}
We report the full results for all variations of NP, RAG, PAG, and RAG+PAG across the baselines in few shot scenario : Table \ref{tab:qwen_fewshot_full_table}, full-data scenario : Table \ref{tab:qwen_fullshot_full_table}. The main results in Table \ref{tab:main-result-qwen-only}, Table \ref{tab:lamp_full_singlecol_avg} correspond to the best-performing configuration among these variations.

\paragraph{Hypernetwork Inference Overhead.}
Text-to-LoRA performs task adaptation purely through inference by using a
hypernetwork that generates LoRA parameters from a natural-language task
description. The hypernetwork used in Text-to-LoRA is the same model
employed in the personalization stage of PRISP; consequently, the
GPU memory required for generating LoRA parameters is limited to the
inference overhead of this network and remains within the peak GPU usage
of the personalization stage. In practice, anchor generation requires
only a single forward pass of the hypernetwork, taking approximately
0.64 seconds with Qwen3-0.6B and 0.69 seconds with Llama-3.1-8B-Instruct,
adding negligible latency to the overall personalization pipeline.
Since the hypernetwork is pre-trained offline and remains fixed during
deployment, Text-to-LoRA incurs virtually no additional cost for new
tasks. Importantly, this step replaces task-data-dependent fine-tuning
rather than introducing an additional training stage, making it a net
reduction in the overall adaptation cost.

\paragraph{Hypernetwork Pre-training Details.}

Following Text-to-LoRA~\citep{charakorn2025texttolora}, we pre-train the hypernetwork using the same training data and procedure as in Text-to-LoRA.
The hypernetwork follows the same architecture as the target backbone model.
For experiments with Qwen-0.6B, we train a Qwen-0.6B-based hypernetwork on the Text-to-LoRA training set, which was conducted on four NVIDIA A100 (80GB) GPUs and took approximately two days.
For experiments with Llama-3.1-8B-Instruct, we directly use the pre-trained hypernetwork provided by Text-to-LoRA\footnote{\url{https://github.com/SakanaAI/text-to-lora}} without additional training.

\paragraph{Training and Inference Setup.}

For each user, we construct training examples from the user's interaction history. Specifically, we generate two versions of each training example: a plain version containing only the target item's input–output pair, and an augmented version that additionally prepends the top-1 BM25-retrieved past interaction as retrieved history context, along with the user's profile text. Both versions are included in the training set for each user, allowing the model to learn from both context-free and retrieval-augmented inputs.

At inference time, the model is evaluated under two prompt settings: (i) without profile (no history or profile text prepended), and (ii) with profile (RAG-retrieved history and user profile text prepended). For the latter, we denote the retrieval strategy as RAG, PAG, or RAG+PAG depending on whether RAG, PAG, or both are applied. The same training and inference protocol is applied consistently across all baselines for fair comparison.

\begin{table}[t]
\caption{Comparison of personalization methods on LaMP benchmarks using Llama-3.1-8B-Instruct.
Arrows indicate whether higher ($\uparrow$) or lower ($\downarrow$) values are better.}
\centering
\footnotesize
\resizebox{\linewidth}{!}{
\begin{tabular}{c l l c c c c}
\toprule
Regime & Task & Metric & Per-Pcs & PriME & OPPU & PRISP \\
\midrule
\multirow{6}{*}{FEW}
 & \multirow{2}{*}{LaMP-1}
 & Acc $\uparrow$   & 0.528 & 0.520 & 0.512 & \textbf{0.536} \\
 &  & F1 $\uparrow$    & \textbf{0.520} & 0.513 & 0.499 & 0.510 \\
\cmidrule(lr){2-7}

 & \multirow{2}{*}{LaMP-2}
 & Acc $\uparrow$   & 0.200 & 0.205 & 0.236 & \textbf{0.491} \\
 &  & F1 $\uparrow$    & 0.156 & 0.161 & 0.198 & \textbf{0.407} \\
\cmidrule(lr){2-7}

 & \multirow{2}{*}{LaMP-4}
 & R-1 $\uparrow$   & 0.169 & 0.171 & 0.188 & \textbf{0.193} \\
 &  & R-L $\uparrow$   & 0.148 & 0.149 & 0.164 & \textbf{0.168} \\

\midrule
\multirow{6}{*}{FULL}
 & \multirow{2}{*}{LaMP-1}
 & Acc $\uparrow$   & \textbf{0.656} & \textbf{0.656} & 0.616 & 0.640 \\
 &  & F1 $\uparrow$    & \textbf{0.652} & \textbf{0.652} & 0.609 & 0.638 \\
\cmidrule(lr){2-7}

 & \multirow{2}{*}{LaMP-2}
 & Acc $\uparrow$   & 0.368 & 0.374 & 0.643 & \textbf{0.679} \\
 &  & F1 $\uparrow$    & 0.341 & 0.350 & 0.561 & \textbf{0.591} \\
\cmidrule(lr){2-7}

 & \multirow{2}{*}{LaMP-4}
 & R-1 $\uparrow$   & 0.191 & 0.189 & \textbf{0.212} & 0.204 \\
 &  & R-L $\uparrow$   & 0.172 & 0.164 & \textbf{0.185} & 0.179 \\

\bottomrule
\label{tab:llama_table}
\end{tabular}
}
\end{table}

\paragraph{Details of Figures.}In Figure~\ref{fig:cost-perform-trade-off}, the x-axis shows the composite cost for each method.
For each task, we first max-normalize GPU memory usage and training time across methods, i.e., the method with the largest GPU memory usage is assigned 1 and all other methods are scaled by their ratio to this maximum; the same normalization is applied to training time.
We then compute the task-level composite cost for each method by averaging the normalized GPU and time values.
Finally, the x-axis value is obtained by averaging these task-level composite costs across all tasks for each method. The y-axis reports task-averaged performance, obtained by first averaging the task-specific evaluation metrics within each task and then averaging across tasks.

\begin{table}[t]
\centering
\caption{Results under aligned data-access settings on LaMP-2 and LaMP-4. 
T and U denote task-level and user-level data access, respectively.}
\label{tab:aligned_eval}
\scriptsize
\setlength{\tabcolsep}{3pt}
\resizebox{0.48\textwidth}{!}{
\begin{tabular}{lcccccc}
\toprule
 & & & \multicolumn{2}{c}{LaMP-2} & \multicolumn{2}{c}{LaMP-4} \\
\cmidrule(lr){4-5} \cmidrule(lr){6-7}
Method & T & U & Acc $\uparrow$ & F1 $\uparrow$ & R-1 $\uparrow$ & R-L $\uparrow$ \\
\midrule
PRISP & \checkmark 50 & 10 & \textbf{0.429} & \textbf{0.304} & \textbf{0.140} & \textbf{0.123} \\
OPPU & \checkmark 50 & 10 & 0.373 & 0.245 & 0.138 & 0.121 \\
RAG+PAG & \checkmark 50 & -- & 0.369 & 0.274 & 0.138 & 0.120 \\
\midrule
PRISP & $\times$ & 10 & \textbf{0.428} & \textbf{0.305} & \textbf{0.138} & \textbf{0.122} \\
OPPU & $\times$ & 10 & 0.358 & 0.224 & 0.135 & 0.117 \\
RAG+PAG & $\times$ & -- & 0.355 & 0.255 & 0.131 & 0.113 \\
\bottomrule
\end{tabular}
}
\end{table}

\begin{table}[t]
\centering
\caption{Results on Product Review and Generation Abstract tasks in LongLaMP benchmark.}
\label{tab:LongLaMP Results}
\small
\resizebox{0.48\textwidth}{!}{
\begin{tabular}{lcccc}
\toprule
 & \multicolumn{2}{c}{Product Review} & \multicolumn{2}{c}{Generation Abstract} \\
\cmidrule(lr){2-3} \cmidrule(lr){4-5}
Method & R-1 $\uparrow$ & R-L $\uparrow$ & R-1 $\uparrow$ & R-L $\uparrow$ \\
\midrule
NP    & 0.1768 & 0.1268 & 0.1829 & 0.1363 \\
RAG   & 0.1671 & 0.1237 & 0.1807 & 0.1193 \\
OPPU  & 0.1801 & 0.1224 & 0.1801 & 0.1236 \\
PRISP & \textbf{0.1855} & \textbf{0.1289} & \textbf{0.2369} & \textbf{0.1643} \\
\bottomrule
\end{tabular}
}
\end{table}

\begin{table*}[t]
\caption{Task descriptions used for generating task-aware LoRA modules via Text-to-LoRA.}

\centering
\small
\setlength{\tabcolsep}{6pt}
\begin{tabular}{p{0.28\textwidth} p{0.68\textwidth}}
\toprule
\textbf{Task} & \textbf{Task Description (used for Text-to-LoRA)} \\
\midrule
Citation Identification &
Identify the most relevant reference for a given paper title. Select between two candidate references [1] and [2]. \\
\midrule
Movie Tagging &
Classify movies into one of these 15 tags based on their descriptions: sci-fi, based on a book, comedy, action, twist ending, dystopia, dark comedy, classic, psychology, fantasy, romance, thought-provoking, social commentary, violence, or true story. \\
\midrule
Product Rating &
What is the score of the following review on a scale of 1 to 5? \\
\midrule
News Headline Generation &
Generate a headline for an article. \\
\midrule
Scholarly Title Generation &
Generate an appropriate title for an academic paper based on the provided abstract. \\
\midrule
Tweet Paraphrasing &
Paraphrase the following text into a tweet. \\
\bottomrule
\end{tabular}

\label{tab:text_to_lora_task_descriptions}
\end{table*}

\begin{table*}[t]
\caption{Variations of task descriptions used for Section \ref{Ablation:robustness to task description variations} in LaMP-2.}

\centering
\small
\setlength{\tabcolsep}{8pt}
\begin{tabular}{p{0.22\textwidth} p{0.74\textwidth}}
\toprule
\textbf{Variation} & \textbf{Task Description} \\
\midrule
Generated Desc. &
For each movie description, determine the most fitting thematic label by understanding its story elements, tone, and central ideas. \\
\midrule
Imprecise Desc. &
Identify movies based on conversations where individuals express their preferences or experiences with certain films. Focus on distinct titles mentioned. \\
\midrule
Ours &
Classify movies into one of these 15 tags based on their descriptions: sci-fi, based on a book, comedy, action, twist ending, dystopia, dark comedy, classic, psychology, fantasy, romance, thought-provoking, social commentary, violence, or true story. \\
\bottomrule
\end{tabular}

\label{tab:movie_tagging_task_description_variations}
\end{table*}

\begin{table*}[t]
\centering
\caption{Main results on LaMP using Qwen3-0.6B in few-shot scenario.}
\resizebox{\textwidth}{!}{
\begin{tabular}{ll c c cc cccc cccc cccc cccc}
\toprule
\multirow{2}{*}{\textbf{Task}} &
\multirow{2}{*}{\textbf{Metric}} &
\textbf{NP} &
\textbf{RAG} &
\multicolumn{2}{c}{\textbf{PAG}} &
\multicolumn{4}{c}{\textbf{Per-Pcs}} &
\multicolumn{4}{c}{\textbf{PriME}} &
\multicolumn{4}{c}{\textbf{OPPU}} &
\multicolumn{4}{c}{\textbf{PRISP}} \\
\cmidrule(lr){3-3}
\cmidrule(lr){4-4}
\cmidrule(lr){5-6}
\cmidrule(lr){7-10}
\cmidrule(lr){11-14}
\cmidrule(lr){15-18}
\cmidrule(lr){19-22}
 &  & $k{=}0$ & $k{=}1$ & $k{=}0$ & $k{=}1$
 & Base & +RAG & +PAG & RAG+PAG
 & Base & +RAG & +PAG & RAG+PAG
 & Base & +RAG & +PAG & RAG+PAG
 & Base & +RAG & +PAG & RAG+PAG \\
\midrule
\multirow{2}{*}{\makecell[l]{LaMP-1}}
& Acc $\uparrow$
& 0.448 & 0.456 & 0.488 & 0.456
& 0.480 & 0.480 & 0.472 & 0.472
& 0.480 & 0.480 & 0.504 & 0.372
& 0.472 & 0.480 & 0.488 & 0.472
& 0.520 & 0.472 & 0.504 & 0.488 \\
& F1 $\uparrow$
& 0.409 & 0.337 & 0.404 & 0.337
& 0.421 & 0.349 & 0.356 & 0.345
& 0.421 & 0.349 & 0.414 & 0.345
& 0.464 & 0.349 & 0.386 & 0.345
& 0.474 & 0.321 & 0.405 & 0.354 \\
\midrule
\multirow{2}{*}{\makecell[l]{LaMP-2}}
& Acc $\uparrow$
& 0.084 & 0.333 & 0.284 & 0.369
& 0.081 & 0.327 & 0.287 & 0.362
& 0.084 & 0.326 & 0.288 & 0.362
& 0.089 & 0.337 & 0.302 & 0.373
& 0.412 & 0.411 & 0.420 & 0.428 \\
& F1 $\uparrow$
& 0.049 & 0.224 & 0.172 & 0.274
& 0.043 & 0.223 & 0.174 & 0.240
& 0.045 & 0.226 & 0.178 & 0.240
& 0.054 & 0.226 & 0.178 & 0.245
& 0.292 & 0.292 & 0.294 & 0.305 \\
\midrule
\multirow{2}{*}{\makecell[l]{LaMP-3}}
& MAE $\downarrow$
& 0.464 & 0.430 & 0.577 & 0.446
& 0.556 & 0.420 & 0.595 & 0.458
& 0.606 & 0.430 & 0.606 & 0.443
& 0.410 & 0.412 & 0.515 & 0.412
& 0.358 & 0.345 & 0.360 & 0.339  \\
& RMSE $\downarrow$
& 0.844 & 0.841 & 1.040 & 0.841
& 1.065 & 0.819 & 1.073 & 0.859
& 1.130 & 0.827 & 1.100 & 0.835
& 0.791 & 0.814 & 0.953 & 0.796
& 0.713 & 0.699 & 0.712 & 0.697 \\
\midrule
\multirow{2}{*}{\makecell[l]{LaMP-4}}
& R-1 $\uparrow$
& 0.132 & 0.129 & 0.135 & 0.138
& 0.128 & 0.121 & 0.134 & 0.133
& 0.128 & 0.122 & 0.135 & 0.133
& 0.123 & 0.122 & 0.135 & 0.138
& 0.131 & 0.134 & 0.133 & 0.138 \\

& R-L $\uparrow$
& 0.114 & 0.113 & 0.116 & 0.120
& 0.111 & 0.107 & 0.116 & 0.117
& 0.110 & 0.107 & 0.116 & 0.117
& 0.106 & 0.108 & 0.117 & 0.121
& 0.115 & 0.118 & 0.117 & 0.122 \\
\midrule
\multirow{2}{*}{\makecell[l]{LaMP-5}}
& R-1 $\uparrow$
& 0.435 & 0.434 & 0.298 & 0.437
& 0.396 & 0.390 & 0.415 & 0.420
& 0.410 & 0.389 & 0.282 & 0.396
& 0.437 & 0.441 & 0.280 & 0.441
& 0.352 & 0.396 & 0.359 & 0.417 \\
& R-L $\uparrow$
& 0.361 & 0.374 & 0.248 & 0.377
& 0.337 & 0.329 & 0.347 & 0.357
& 0.340 & 0.325 & 0.235 & 0.337
& 0.363 & 0.379 & 0.232 & 0.377
& 0.313 & 0.340 & 0.310 & 0.363 \\
\midrule
\multirow{2}{*}{\makecell[l]{LaMP-7}}
& R-1 $\uparrow$
& 0.441 & 0.477 & 0.381 & 0.399
& 0.222 & 0.165 & 0.171 & 0.151
& 0.325 & 0.311 & 0.281 & 0.281
& 0.429 & 0.437 & 0.368 & 0.397
& 0.356 & 0.420 & 0.378 & 0.394 \\
& R-L $\uparrow$
& 0.395 & 0.397 & 0.346 & 0.362
& 0.208 & 0.154 & 0.158 & 0.144
& 0.288 & 0.285 & 0.247 & 0.257
& 0.382 & 0.391 & 0.331 & 0.360
& 0.322 & 0.382 & 0.337 & 0.352 \\
\bottomrule
\end{tabular}
}
\label{tab:qwen_fewshot_full_table}
\end{table*}

\begin{table*}[t]
\centering
\caption{Main results on LaMP using Qwen3-0.6B in the full-data scenario.}
\resizebox{\textwidth}{!}{
\begin{tabular}{ll c c cc cccc cccc cccc cccc}
\toprule
\multirow{2}{*}{\textbf{Task}} &
\multirow{2}{*}{\textbf{Metric}} &
\textbf{NP} &
\textbf{RAG} &
\multicolumn{2}{c}{\textbf{PAG}} &
\multicolumn{4}{c}{\textbf{Per-Pcs}} &
\multicolumn{4}{c}{\textbf{PriME}} &
\multicolumn{4}{c}{\textbf{OPPU}} &
\multicolumn{4}{c}{\textbf{PRISP}} \\
\cmidrule(lr){3-3}
\cmidrule(lr){4-4}
\cmidrule(lr){5-6}
\cmidrule(lr){7-10}
\cmidrule(lr){11-14}
\cmidrule(lr){15-18}
\cmidrule(lr){19-22}
 &  & $k{=}0$ & $k{=}1$ & $k{=}0$ & $k{=}1$
 & Base & +RAG & +PAG & RAG+PAG
 & Base & +RAG & +PAG & RAG+PAG
 & Base & +RAG & +PAG & RAG+PAG
 & Base & +RAG & +PAG & RAG+PAG \\
\midrule
\multirow{2}{*}{\makecell[l]{LaMP-1}}
& Acc $\uparrow$
& 0.600 & 0.584 & 0.704 & 0.696
& 0.560 & 0.592 & 0.704 & 0.696
& 0.600 & 0.608 & 0.712 & 0.728
& 0.600 & 0.592 & 0.736 & 0.736
& 0.544 & 0.504 & 0.664 & 0.704 \\
& F1 $\uparrow$
& 0.597 & 0.584 & 0.699 & 0.693
& 0.525 & 0.592 & 0.691 & 0.694
& 0.570 & 0.608 & 0.702 & 0.726
& 0.570 & 0.592 & 0.733 & 0.734
& 0.541 & 0.497 & 0.663 & 0.704 \\
\midrule
\multirow{2}{*}{\makecell[l]{LaMP-2}}
& Acc $\uparrow$
& 0.256 & 0.435 & 0.432 & 0.496
& 0.251 & 0.432 & 0.432 & 0.505
& 0.249 & 0.429 & 0.436 & 0.498
& 0.442 & 0.478 & 0.491 & 0.518
& 0.574 & 0.570 & 0.578 & 0.580 \\
& F1 $\uparrow$
& 0.220 & 0.345 & 0.345 & 0.393
& 0.208 & 0.340 & 0.344 & 0.403
& 0.214 & 0.341 & 0.348 & 0.391
& 0.306 & 0.363 & 0.375 & 0.403
& 0.468 & 0.463 & 0.464 & 0.477 \\
\midrule
\multirow{2}{*}{\makecell[l]{LaMP-3}}
& MAE $\downarrow$
& 0.329 & 0.282 & 0.313 & 0.283
& 0.453 & 0.349 & 0.368 & 0.331
& 0.446 & 0.355 & 0.375 & 0.324
& 0.287 & 0.272 & 0.283 & 0.265
& 0.292 & 0.279 & 0.282 & 0.269 \\
& RMSE $\downarrow$
& 0.690 & 0.589 & 0.651 & 0.604
& 0.862 & 0.690 & 0.724 & 0.634
& 0.869 & 0.704 & 0.733 & 0.627
& 0.636 & 0.613 & 0.612 & 0.600
& 0.611 & 0.592 & 0.608 & 0.578 \\
\midrule
\multirow{2}{*}{\makecell[l]{LaMP-4}}
& R-1 $\uparrow$
& 0.143 & 0.153 & 0.148 & 0.157
& 0.132 & 0.143 & 0.136 & 0.148
& 0.150 & 0.143 & 0.153 & 0.149
& 0.150 & 0.156 & 0.155 & 0.160
& 0.146 & 0.154 & 0.152 & 0.156 \\
& R-L $\uparrow$
& 0.123 & 0.133 & 0.129 & 0.137
& 0.114 & 0.125 & 0.118 & 0.130
& 0.132 & 0.125 & 0.133 & 0.130
& 0.131 & 0.137 & 0.135 & 0.140
& 0.128 & 0.135 & 0.134 & 0.137 \\
\midrule
\multirow{2}{*}{\makecell[l]{LaMP-5}}
& R-1 $\uparrow$
& 0.412 & 0.437 & 0.420 & 0.440
& 0.412 & 0.422 & 0.415 & 0.421
& 0.411 & 0.422 & 0.411 & 0.422
& 0.412 & 0.435 & 0.423 & 0.444
& 0.424 & 0.434 & 0.427 & 0.433 \\
& R-L $\uparrow$
& 0.356 & 0.380 & 0.363 & 0.382
& 0.355 & 0.371 & 0.358 & 0.368
& 0.353 & 0.369 & 0.356 & 0.370
& 0.351 & 0.381 & 0.365 & 0.384
& 0.366 & 0.381 & 0.370 & 0.379 \\
\midrule
\multirow{2}{*}{\makecell[l]{LaMP-7}}
& R-1 $\uparrow$
& 0.455 & 0.469 & 0.459 & 0.489
& 0.459 & 0.458 & 0.454 & 0.463
& 0.463 & 0.460 & 0.467 & 0.467
& 0.455 & 0.477 & 0.455 & 0.474
& 0.462 & 0.473 & 0.472 & 0.495 \\
& R-L $\uparrow$
& 0.407 & 0.429 & 0.406 & 0.448
& 0.410 & 0.421 & 0.408 & 0.424
& 0.411 & 0.423 & 0.423 & 0.426
& 0.402 & 0.430 & 0.405 & 0.435
& 0.408 & 0.427 & 0.413 & 0.445 \\
\bottomrule
\end{tabular}
}
\label{tab:qwen_fullshot_full_table}
\end{table*}

\begin{figure}[t]
    \centering
    \includegraphics[width=\linewidth]{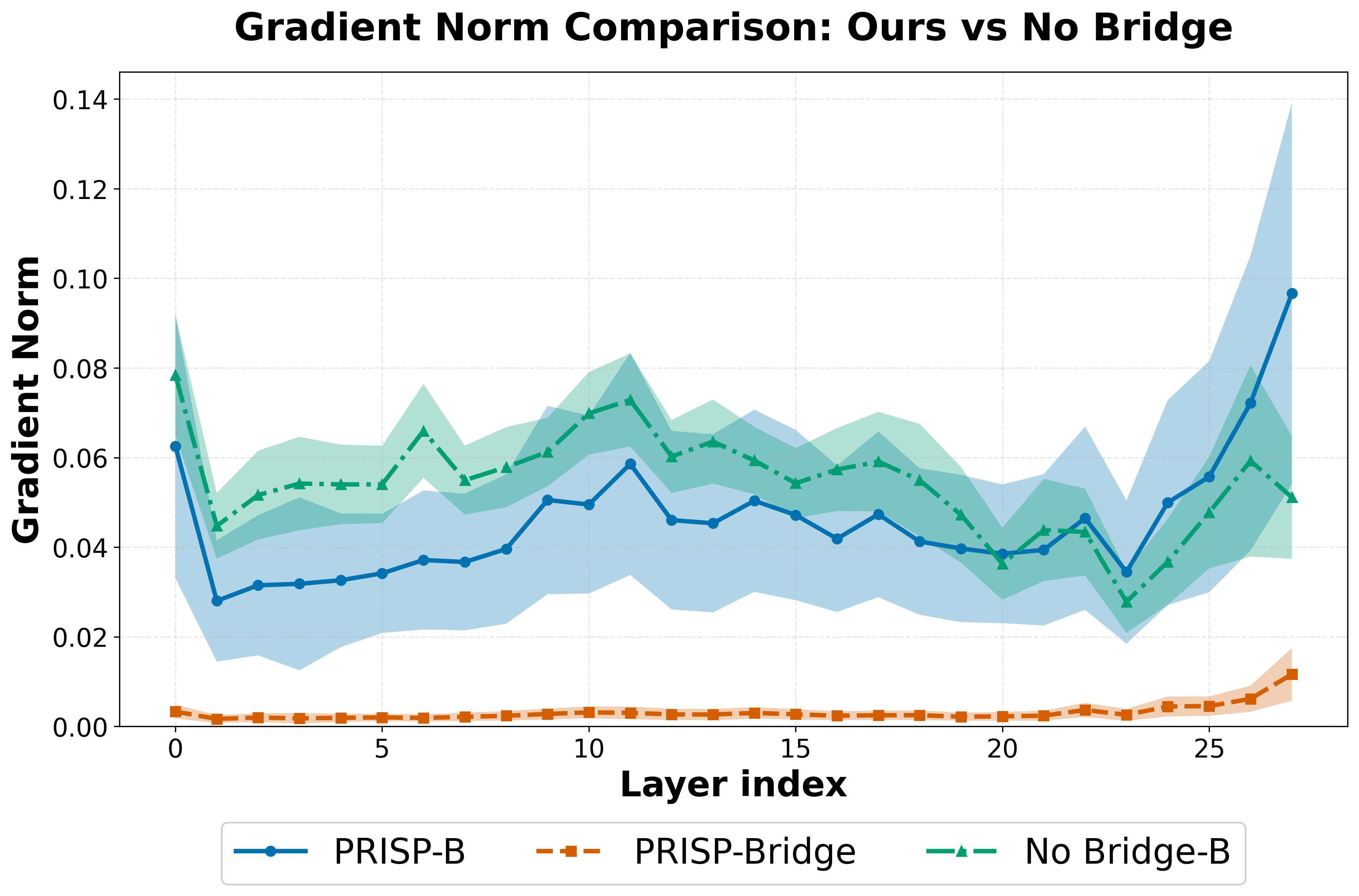}
    \caption{Comparison of gradient norms between PRISP and No-Bridge variant during training.
Gradient norms are computed in a layer-wise manner and averaged across all Transformer layers at each training step.
Since training is performed with a single epoch in few-shot setting, we increase the number of optimization steps for clearer observation by setting the batch size to 1.}
    \label{fig:gradient-norm}
\end{figure}

\section{Additional Analyses}
\label{appendix:C_addditional Analyses}

\paragraph{Aligned Data setting Comparison.}

We conducted additional experiments under fully aligned data-access settings. Specifically, (i) we allow PRISP to use 50 task-level samples during personalization, matching the data access of task-adaptive baselines and (ii) we evaluate OPPU, RAG+PAG without access to task-level data, using only user history, to match our task-data-free setting.

As shown in below table \ref{tab:aligned_eval}, when PRISP is additionally provided with 50 task samples, its performance shows a slight improvement, confirming that task-level supervision can offer some benefit. However, the gains remain modest, indicating that PRISP already performs strongly even without task data. In contrast, OPPU and RAG+PAG experience substantial drops when task-level data are removed, demonstrating that their effectiveness relies heavily on task supervision.

These results suggest that our few-shot protocol is, if anything, more favorable to baseline methods that depend on task-level data. PRISP is evaluated under strictly harsher conditions, reinforcing the robustness of our findings.

\paragraph{Scaling Behavior on Larger Models.}

We further analyze PRISP on a larger backbone, Llama-3.1-8B-Instruct \citep{grattafiori2024llama3}. Table~\ref{tab:llama_table} reports results under both few-shot and full-data personalization settings.
Under the few-shot regime, PRISP consistently delivers strong performance across all evaluated tasks. In the full-data regime, it remains competitive with competing methods, 
indicating that our approach does not trade off performance in data-rich settings while offering clear advantages under data-scarce personalization.
Overall, these results suggest that PRISP scales favorably with model size, maintaining its effectiveness on larger-capacity backbones.

\paragraph{Generalization to Long-Context Personalization.}

To further evaluate generalization beyond the original task set, we conduct additional experiments on LongLaMP \cite{kumar2024longlamp}, a long-context personalization benchmark for open-ended generation tasks that requires models to utilize extended user history. The benchmark covers diverse long-form generation tasks, including abstract generation and product review writing. We follow the same few-shot protocol as in our main experiments, using 50 task-level samples and 10 user-level samples per user. The results are summarized in Table \ref{tab:LongLaMP Results}.

\paragraph{Gradient Analysis of Bridge Matrix.}

To better understand the optimization dynamics of PRISP in Section \ref{sec:ablation studies}, we analyze the gradient norms of LoRA parameters during training, focusing on the LoRA $B$ matrices, which constitute the primary trainable components in both PRISP and the No-Bridge variant.

In Figure \ref{fig:gradient-norm}, We observe that early layers (e.g., layers 3–11), which typically capture more general features, show comparable gradient magnitudes across variants. In contrast, later layers (e.g., layers 19–27), which are more responsible for user-specific representations, exhibit significantly larger and more structured gradients under the bridge formulation compared to the No-Bridge variant. We attribute this behavior to the joint optimization of the LoRA $B$ matrix and the bridge matrix in PRISP, as opposed to the No-Bridge variant where the $B$ matrix is optimized alone.

This indicates that the bridge matrix selectively amplifies adaptation in deeper layers, enabling more effective user-level personalization without destabilizing shared representations.

\section{Theoretical Justification of Stage-2 Design}
\label{appendix:theoretical_justification}

In this section, we provide formal justifications for the design choices in Stage-2 of PRISP, focusing on (1) space complexity and memory efficiency, and (2) why freezing the anchor improves few-shot robustness.

\subsection{Space Complexity and Memory Efficiency}

For a layer with input dimension $d_{\text{in}}$, output dimension $d_{\text{out}}$, and rank $r$:

\paragraph{Standard LoRA.} The weight update is parameterized as $\Delta W = BA$, where $A \in \mathbb{R}^{r \times d_{\text{in}}}$ and $B \in \mathbb{R}^{d_{\text{out}} \times r}$. The total number of trainable parameters is:
\begin{equation*}
    |\theta_{\text{LoRA}}| = r(d_{\text{in}} + d_{\text{out}}).
\end{equation*}

\paragraph{PRISP (Stage-2).} The weight update is parameterized as $\Delta W = B_t C A_t$, where $A_t$ is frozen, and we train $B_t \in \mathbb{R}^{d_{\text{out}} \times r}$ and $C \in \mathbb{R}^{r \times r}$. The total number of trainable parameters becomes:
\begin{equation*}
    |\theta_{\text{PRISP}}| = d_{\text{out}} r + r^2.
\end{equation*}

The ratio of trainable parameters is:
\begin{equation*}
    \frac{|\theta_{\text{PRISP}}|}{|\theta_{\text{LoRA}}|} = \frac{d_{\text{out}} + r}{d_{\text{in}} + d_{\text{out}}}.
\end{equation*}

For typical transformer layers where $d_{\text{in}} = d_{\text{out}} = d$ and $r \ll d$, this ratio is approximately $1/2$, i.e., about 50\% fewer trainable parameters.

Since training memory is dominated by gradients and optimizer states, freezing $A_t$ removes its gradient and optimizer tensors entirely, leading to proportional memory reduction.

\paragraph{Stage-2 Parameter Efficiency.}
During personalization, we optimize only $B_t$ and a small bridge matrix $C \in \mathbb{R}^{r \times r}$ (with $r = 8$) while freezing $A_t$. This results in roughly half the number of trainable parameters ($\sim$50\%) compared to standard LoRA fine-tuning. The bridge adds only $r^2$ parameters per layer (64 when $r = 8$), which is negligible relative to backbone size.

\subsection{Why Freezing the Anchor Improves Few-Shot Robustness}

In extreme few-shot settings, overfitting stems from high-variance updates. A standard way to mitigate this under limited data is to reduce the number of trainable degrees of freedom, i.e., optimize fewer parameters, thereby lowering model capacity. However, merely reducing parameter count is not sufficient; what matters is restricting adaptation to a well-structured subspace aligned with the task prior.

In PRISP, the adaptation is parameterized as $\Delta W = B_t C A_t$, and freezing $A_t$ restricts learning to variations in $B_t$ and $C$ only. This restricts learning to a task-informed low-dimensional subspace, acting as a strong structural prior and reducing variance:
\begin{equation*}
\begin{split}
    \Delta W &= B_t C A_t, \\
    A_t &\in \mathbb{R}^{r \times d_{\text{in}}} \; (\text{frozen}), \; B_t \in \mathbb{R}^{d_{\text{out}} \times r}, \; C \in \mathbb{R}^{r \times r}.
\end{split}
\end{equation*}

Because $A_t$ is fixed, any update satisfies the rank and subspace constraint:
\begin{equation*}
    \text{rank}(\Delta W) \leq r, \qquad \text{Im}(\Delta W^\top) \subseteq \text{Im}(A_t^\top).
\end{equation*}

Indeed, since $\Delta W^\top = A_t^\top C^\top B_t^\top$, every column of $\Delta W^\top$ lies in the column space of $A_t^\top$.

Equivalently, the update acts only through the $r$-dimensional projected features:
\begin{equation*}
    \Delta W \cdot h = B_t C (A_t h), \quad z := A_t h \in \mathbb{R}^r,
\end{equation*}
so personalization is restricted to a task-aligned subspace determined by the anchor.

Thus the effective hypothesis class becomes:
\begin{equation*}
    \mathcal{H}_{\text{PRISP}} = \left\{ W_0 + B C A_t \;\middle|\; \begin{array}{l} B \in \mathbb{R}^{d_{\text{out}} \times r}, \\ C \in \mathbb{R}^{r \times r} \end{array} \right\},
\end{equation*}
which is a strict restriction compared to optimizing both LoRA factors. This capacity restriction acts as a structural prior and reduces estimation variance in the extreme few-shot regime, improving robustness against overfitting and bias.

In addition, the bridge matrix ($C \in \mathbb{R}^{r \times r}$) restores controlled flexibility within this subspace, preventing underfitting. Empirically, this outperforms the ``No-Bridge'' variant (training only $B$), indicating a better bias--variance trade-off (Table~\ref{tab:Comparison of bridge}).

To further support this, we provide a layer-wise gradient analysis of the bridge mechanism (Appendix~\ref{appendix:C_addditional Analyses}, Figure~\ref{fig:gradient-norm}).

% \begin{table}[t]
% \centering
% \small
% \setlength{\tabcolsep}{6pt}
% \begin{tabular}{p{0.22\textwidth} p{0.72\textwidth}}
% \toprule
% \textbf{Task} & \textbf{Task Description} \\
% \midrule
% Citation Identification &
% You are given the title of an academic paper and must determine which reference is more closely related to its topic by understanding the core ideas of both the title and the reference summaries. \\
% \midrule
% Movie Tagging &
% For each movie description, determine the most fitting thematic label by understanding its story elements, tone, and central ideas. \\
% \midrule
% Product Rating &
% You read a product review and decide how positive or negative the customer’s experience was by assigning a score that reflects their satisfaction. \\
% \midrule
% News Headline Generation &
% The task requires reading a full news article and summarizing its core idea into a concise headline that captures the main message. \\
% \midrule
% Scholarly Title Generation &
% Your goal is to understand the problem, methods, and contributions described in a research abstract, then summarize them in the form of an academic paper title. \\
% \midrule
% Tweet Paraphrasing &
% You are given a piece of text and must rewrite it so that it fits the tone, style, and brevity commonly found in tweets. \\
% \bottomrule
% \end{tabular}
% \caption{High-level task descriptions used for task description generation and text-to-LoRA conditioning.}
% \label{tab:task_descriptions}
% \end{table}

\begin{table*}[t]
\centering
\small
\setlength{\tabcolsep}{8pt}
\renewcommand{\arraystretch}{1.15}
\caption{Profile generation prompts used in our experiments.}
\label{tab:appendix_profile_prompts}
\begin{tabularx}{\textwidth}{@{}>{\bfseries}l >{\RaggedRight\arraybackslash}X@{}}
\toprule
Task & Prompt Template \\
\midrule
LaMP-1 &
Write a summary, in English, of the research interests and topics of a researcher who has published the following papers.
Only generate the summary, no other text.
User history: \{USER HISTORY\}
Answer: \\
\midrule
LaMP-2 &
Look at the following past movies this user has watched and determine the most popular tag they labeled.
Answer in the following form: most popular tag: \texttt{<tag>}.
User History: \{USER HISTORY\}
Answer: \\
\midrule
LaMP-3 &
Based on this user’s past reviews, what are the most common scores they give for positive and negative reviews?
Answer in the following form: most common positive score: \texttt{<most common positive score>}, most common negative score: \texttt{<most common negative score>}.
User History: \{USER HISTORY\}
Answer: \\
\midrule
LaMP-4 &
Given this author’s previous articles, try to describe a template for their headlines.
I want to be able to accurately predict the headline given one of their articles.
Be specific about their style and wording, don’t tell me anything generic.
User History: \{USER HISTORY\}
Answer: \\
\midrule
LaMP-5 &
Given this author’s previous publications, try to describe a template for their titles.
I want to be able to accurately predict the title of one of the papers from the abstract.
Only generate the template description, nothing else.
User History: \{USER HISTORY\}
Answer: \\
\midrule
LaMP-7 &
Given this person’s previous tweets, try to describe a template for their tweets.
I want to take a generic sentence and rephrase it to sound like one of their tweets, with the same style/punctuation/capitalization/wording/tone/etc. as them.
Only give me the template description, nothing else.
User History: \{USER HISTORY\}
Answer: \\
\bottomrule
\end{tabularx}
\label{tab: profile generation}
\end{table*}

\begin{table*}[t]

\centering
\small
\setlength{\tabcolsep}{6pt}
\caption{Personalization prompts. Prompt templates used for personalization.}
\begin{tabular}{p{0.12\textwidth} p{0.83\textwidth}}
\toprule
\textbf{Task} & \textbf{Prompt Template} \\
\midrule
\textbf{LaMP-1} &
\textbf{User Instruction:}\\
& Identify the most relevant reference for the listed publication by the researcher.
Select the reference paper that is most closely related to the researcher’s work.
Please respond with only the number that corresponds to the reference. \\
& Paper Title: \{QUERY PAPER TITLE\} \\
& Reference: [1] - \{OPTION1\} [2] - \{OPTION2\} \\
& Answer: \\
\midrule
\textbf{LaMP-2} &
\textbf{User Instruction:}\\
& Which tag does this movie relate to among the following tags?
Just answer with the tag name without further explanation. \\
& tags: [sci-fi, based on a book, comedy, action, twist ending, dystopia, dark comedy,
classic, psychology, fantasy, romance, thought-provoking, social commentary,
violence, true story] \\
& Description: \{QUERY MOVIE DESCRIPTION\} \\
& Tag: \\
\midrule
\textbf{LaMP-3} &
\textbf{User Instruction:}\\
& What is the score of the following review on a scale of 1 to 5?
Just answer with 1, 2, 3, 4, or 5 without further explanation. \\
& Review: \{QUERY REVIEW\} \\
& Score: \\
\midrule
\textbf{LaMP-4} &
\textbf{User Instruction:}\\
& Generate a headline for the following article. \\
& Article: \{QUERY ARTICLE\} \\
& Headline: \\
\midrule
\textbf{LaMP-5} &
\textbf{User Instruction:}\\
& Generate a title for the following abstract of a paper. \\
& Abstract: \{QUERY ABSTRACT\} \\
& Title: \\
\midrule
\textbf{LaMP-7} &
\textbf{User Instruction:}\\
& Paraphrase the following text into tweet without any explanation before or after it. \\
& Text: \{QUERY TEXT\} \\
& Tweet: \\
\bottomrule
\end{tabular}

\label{tab:personalization_prompts}
\end{table*}

\end{document}